\documentclass[12pt,eqsecnum,aps,preprintnumbers,groupedaddress,prd,nofootinbib]{revtex4}
\usepackage{amsmath,amsfonts,bm}
\usepackage{hyperref}
\usepackage{url}

\DeclareMathAlphabet{\mathsfit}{\encodingdefault}{\sfdefault}{m}{sl}
\SetMathAlphabet{\mathsfit}{bold}{\encodingdefault}{\sfdefault}{bx}{n}

\usepackage[pdftex]{graphicx}

\def\le{\left}
\def\ri{\right}
\newcommand{\be}{\begin{equation}}
\newcommand{\ee}{\end{equation}}
\newcommand{\bea}{\begin{eqnarray}}
\newcommand{\eea}{\end{eqnarray}}

\def\ND{N_{\text{D}}} 
\def\NR{N_{\text{R}}} 
\def\NE{N_{\text{E}}} 
\def\tra{\bar{\beta}}
\def\tes{\dot{\gamma}}

\def\ker{K} 
\def\kert{\widetilde{K}} 
\def\ac{\mathcal{H}} 
\def\PCF{G} 
\def\ACF{H} 
\def\PCFt{\bm{G}} 
\def\ACFt{\bm{H}} 
\def\FPV{\widetilde{V}} 
\def\FPVA{V} 
\def\SE{\widetilde{S}} 

\def\inp{\bm{x}} 
\def\outp{\bm{z}} 
\def\rvz{\mathbf{z}} 
\def\rz{\textnormal{z}} 
\def\MP{\bm{\theta}} 
\def\tar{y} 
\def\GPM{\textnormal{y}^{\text{GP}}_{\text{E}}} 
\def\YC{\bm{y}_{\text{R}}} 
\def\YT{\overline{\textnormal{y}}_{\text{E}}} 

\begin{document}

\title{Non-Gaussian processes and neural networks at finite widths}

\author{Sho Yaida\\
Facebook AI Research\\
Facebook Inc.\\
Menlo Park, California 94025, USA \\
\texttt{shoyaida@fb.com}
}

\begin{abstract}
Gaussian processes are ubiquitous in nature and engineering. A case in point is a class of neural networks in the infinite-width limit, whose priors correspond to Gaussian processes. Here we perturbatively extend this correspondence to finite-width neural networks, yielding non-Gaussian processes as priors. The methodology developed herein allows us to track the flow of preactivation distributions by progressively integrating out random variables from lower to higher layers, reminiscent of renormalization-group flow. We further develop a perturbative procedure to perform Bayesian inference with weakly non-Gaussian priors.
\end{abstract}

\maketitle

\section{Inception}\label{introduction}
Gaussian processes model many phenomena in the physical world. A prime example is Brownian motion~\cite{Brown1828}, modeled as the integral of Gaussian-distributed bumps exerted on a point-like solute~\cite{Einstein1905}. The theory of elementary particles~\cite{Weinberg1995} also becomes a Gaussian process in the free limit where interactions between particles are turned off, and many-body systems as complex as glasses come to be Gaussian in the infinite-dimensional, mean-field, limit~\cite{PZ2010}. In the context of machine learning, Neal in Ref.~\cite{Neal1996} pointed out that a class of neural networks give rise to Gaussian processes in the infinite-width limit, which can perform exact Bayesian inference from training to test data~\cite{Williams1997}. They occupy a corner of theoretical playground wherein the \textit{karakuri} of neural networks is scrutinized~\cite{LBNSPS2017,MRHTG2018,JGH2018,COB2019,LXSBSP2019,GSJW2019}.

In reality, Gaussian processes are but mere idealizations. Brownian particles have finite-size structure, elementary particles interact, and many-body systems respond nonlinearly. In order to understand rich phenomena exhibited by these real systems, Gaussian processes rather serve as starting points to be perturbed around. Indeed many edifices in theoretical physics are built upon the successful treatment of non-Gaussianity, with a notable example being renormalization-group flow~\cite{Kadanoff1966,Wilson1971,Weinberg1996,Goldenfeld2018}. In the quest to elucidate behaviors of real neural networks away from the infinite-width limit, it is thus natural to wonder if the similar treatment of non-Gaussianity yields equally elegant and powerful machinery.

Here we set out on this program, perturbatively treating finite-width corrections to neural networks. Prior distributions of outputs are obtained through progressively integrating out preactivations of neurons layer by layer, yielding non-Gaussian priors. The whole procedure closely resembles renormalization-group flow~\cite{Goldenfeld2018,MS2014}: it bridges probability distributions at different scales through coarse-graining of random variables at microscopic scales; the flow of distributions is traced through running couplings, which in particular capture the degree of non-Gaussianity in these distributions; resulting recursive equations~(\ref{R1},\ref{R2},\ref{R3}) govern the evolution of these running couplings from lower to higher layers, just as renormalization-group equations do from microscopic to macroscopic scales. Such a recursive approach enables us to treat finite-width corrections to various observables, for networks with arbitrary activation functions.

The rest of the paper is structured as follows. In Section~\ref{setup} we review and set up basic concepts. Our master recursive formulae~(\ref{R1},\ref{R2},\ref{R3}) are derived in Section~\ref{flow}, which control the flow of preactivation distributions. After an interlude with concrete examples in Section~\ref{examples}, we extend the Gaussian-process Bayesian inference to non-Gaussian priors in Section~\ref{Bayes} and study inference of neural networks at finite widths. We conclude in Section~\ref{conclusion} with dreams.

\section{To infinity and beyond}\label{setup}
In this paper we study real finite-width neural networks in the regime where the number of neurons in hidden layers is asymptotically large whereas input and output dimensions are kept constant.

\subsection{Gaussian processes and neural networks at infinite widths}
Let us focus on a class of neural networks termed multilayer perceptrons, with model parameters, $\MP=\le\{b^{(\ell)}_{i},W^{(\ell)}_{ij}\ri\}$, and an activation function, $\sigma$. For each input, $\inp\in \mathbb{R}^{n_{0}}$, a neural network outputs a vector, $\outp(\inp; \MP)=\outp^{(L)}\in \mathbb{R}^{n_{L}}$, recursively defined as sequences of preactivations through
\bea
z^{(1)}_{i}(\inp)&=&b^{(1)}_{i}+\sum_{j=1}^{n_{0}}W^{(1)}_{ij}x_{j} \ \ \ \text{for}\ \ \ i=1,\ldots,n_{1}\, ,\\
z^{(\ell)}_{i}(\inp)&=&b^{(\ell)}_{i}+\sum_{j=1}^{n_{\ell-1}}W^{(\ell)}_{ij}\sigma\le[z^{(\ell-1)}_{j}(\inp)\ri] \ \ \ \text{for}\ \ \ i=1,\ldots,n_{\ell}\, ; \ \ \ \ell=2,\ldots,L\, .
\eea
Following Ref.~\cite{Neal1996}, we assume priors for biases and weights given by independent and identically distributed Gaussian distributions with zero means, $\mathbb{E}\le[b^{(\ell)}_i\ri]=\mathbb{E}\le[W^{(\ell)}_{ij}\ri]=0$, and variances
\bea
\mathbb{E}\le[b^{(\ell)}_{i_1}b^{(\ell)}_{i_2}\ri]&=&\delta_{i_1 i_2} C_{b}^{(\ell)}\, ,\\
\mathbb{E}\le[W^{(\ell)}_{i_1 j_1}W^{(\ell)}_{i_2 j_2}\ri]&=&\delta_{i_1 i_2} \delta_{j_1 j_2}\frac{C_{W}^{(\ell)}}{n_{\ell-1}}\, .
\eea
Higher moments are then obtained by Wick's contractions~\cite{Wick1950,Zee2010}. For instance,
\be
\mathbb{E}\le[b^{(\ell)}_{i_1}b^{(\ell)}_{i_2}b^{(\ell)}_{i_3}b^{(\ell)}_{i_4}\ri]=\le[C_{b}^{(\ell)}\ri]^2\times\le(\delta_{i_1 i_2}\delta_{i_3 i_4}+\delta_{i_1 i_3}\delta_{i_2 i_4} +\delta_{i_1 i_4}\delta_{i_2 i_3} \ri)\, .
\ee
For those unfamiliar with Wick's contractions and connected correlation functions (a.k.a.~cumulants), a pedagogical review is provided in Appendix~\ref{cumulants_for_dummies} as our formalism heavily relies on them.

In the infinite-width limit where $n_{1},n_{2},\ldots,n_{L-1}\rightarrow\infty$ (but finite $n_0$ and $n_L$), it has been argued -- with varying degrees of rigor~\cite{Neal1996,LBNSPS2017,MRHTG2018} -- that the prior distribution of outputs is governed by the Gaussian process with a kernel
\be
\ker_{i_1 i_2; \alpha_1 \alpha_2}\equiv \mathbb{E}\le[z^{(L)}_{i_1}(\inp_{\alpha_1})z^{(L)}_{i_2}(\inp_{\alpha_2})\ri]\, 
\ee 
and all the higher moments given by Wick's contractions. Here, the sample index $\alpha$ labels different inputs in a dataset. There exists a recursive formula that lets us evaluate this kernel for any pair of inputs~\cite{LBNSPS2017}~[c.f.~Equation~(\ref{R1})]. Importantly, once the values of the kernel are evaluated for all the pairs of $\ND=\NR+\NE$ input data, $\le\{\inp_{\alpha}\ri\}_{\alpha=1,\ldots,\ND}$, consisting of $\NR$ training inputs with target outputs and $\NE$ test inputs with unknown targets, we can perform exact Bayesian inference to yield mean outputs as predictions for $\NE$ test data~\cite{Williams1997,WR2006} [c.f.~Equation~(\ref{GPP})]. This should be contrasted with stochastic gradient descent (SGD) optimization~\cite{RM1951}, through which typically a single estimate for the optimal model parameters of the posterior, $\MP_{\star}$, is obtained and used to predict outputs for test inputs; Bayesian inference instead marginalizes over all model parameters, performing an ensemble average over the posterior distribution~\cite{Mackay1995}.

\subsection{Beyond infinity}
We shall now study real finite-width neural networks in the regime $n_{1},\ldots,n_{L-1}\sim n\gg 1$.\footnote{Note that input and output dimensions, $n_0$ and $n_L$, are arbitrary. To be precise, defining $n_{1},\ldots,n_{L-1}\equiv r_1 n,\ldots,r_{L-1} n$, we send $n\gg1$ while keeping $\le\{C_{b}^{(\ell)},C_{W}^{(\ell)}\ri\}_{\ell=1,\ldots,L}$, $r_1,\ldots,r_{L-1}$, $n_{0}$, and $n_{L}$ constants, and compute the leading $1/n$ corrections. In particular it is crucial to keep the number of outputs $n_{L}$ constant in order to consistently perform Bayesian inference within our approach.}
At finite widths, there are corrections to Gaussian-process priors. In other words, a whole tower of nontrivial preactivation correlation functions beyond the kernel,
\be
\PCF^{(\ell)}_{i_1 \ldots i_m; \alpha_1 \ldots \alpha_m}\equiv\mathbb{E}\le[z^{(\ell)}_{i_1}(\inp_{\alpha_1})\cdots z^{(\ell)}_{i_m}(\inp_{\alpha_m})\ri]\, ,
\ee
collectively dictate the distribution of preactivations. Our aim is to trace the flow of these distributions progressively and cumulatively all the way up to the last layer whereat Bayesian inference is executed. More specifically, we shall inductively and self-consistently show that two-point preactivation correlation functions take the form\footnote{In the main text we place tildes on objects that depend only on sample indices $\alpha$'s in order to distinguish them from those that depend both on sample indices $\alpha$'s and neuron indices $i$'s.}
\be\label{twopoint}\tag{KS}
\PCF^{(\ell)}_{i_1i_2; \alpha_1\alpha_2}=\delta_{i_1 i_2} \le[\kert_{\alpha_1 \alpha_2}^{(\ell)}+\frac{1}{n_{\ell-1}}\SE^{(\ell)}_{\alpha_1\alpha_2}+O\le(\frac{1}{n^{2}}\ri)\ri]\, 
\ee
and connected four-point preactivation correlation functions
\begin{align}\label{fourpoint}\tag{V}
&\PCF^{(\ell)}_{i_1 i_2 i_3 i_4; \alpha_1 \alpha_2 \alpha_3 \alpha_4}\Big|_{\text{connected}}\, \\
\equiv& \PCF^{(\ell)}_{i_1 i_2 i_3 i_4; \alpha_1 \alpha_2 \alpha_3 \alpha_4}-\PCF^{(\ell)}_{i_1 i_2; \alpha_1\alpha_2}\PCF^{(\ell)}_{i_3 i_4; \alpha_3 \alpha_4}-\PCF^{(\ell)}_{i_1 i_3; \alpha_1 \alpha_3}\PCF^{(\ell)}_{i_2 i_4; \alpha_2\alpha_4}-\PCF^{(\ell)}_{i_1 i_4; \alpha_1 \alpha_4}\PCF^{(\ell)}_{i_2 i_3; \alpha_2 \alpha_3}\, \nonumber\\
=&\frac{1}{n_{\ell-1}} \Big[\delta_{i_1i_2}\delta_{i_3i_4} \FPV^{(\ell)}_{(\alpha_1\alpha_2)(\alpha_3,\alpha_4)}+\delta_{i_1 i_3}\delta_{i_2 i_4} \FPV^{(\ell)}_{(\alpha_1\alpha_3)(\alpha_2\alpha_4)}+\delta_{i_1 i_4}\delta_{i_2 i_3} \FPV^{(\ell)}_{(\alpha_1 \alpha_4) (\alpha_2 \alpha_3)}\Big]+O\le(\frac{1}{n^{2}}\ri)\, , \nonumber
\end{align}
and higher cumulants are all suppressed by $O\le(\frac{1}{n^{2}}\ri)$.\footnote{Given that the means of biases and weights are zero, $\PCF^{(\ell)}_{i_1 \ldots i_m; \alpha_1 \ldots \alpha_m}=0$ for all odd $m$.} Here the Gaussian-process core kernel $\kert^{(\ell)}_{\alpha_1 \alpha_2}$ and the self-energy correction $\SE^{(\ell)}_{\alpha_1\alpha_2}$ are symmetric under the exchange of sample indices $\alpha_1\leftrightarrow\alpha_2$ and the four-point vertex $\FPV^{(\ell)}_{(\alpha_1 \alpha_2) (\alpha_3 \alpha_4)}$ is symmetric under $\alpha_1\leftrightarrow\alpha_2$, $\alpha_3\leftrightarrow\alpha_4$, and $(\alpha_1\alpha_2)\leftrightarrow(\alpha_3\alpha_4)$. At the first layer the preactivation distribution is exactly Gaussian for any finite widths and hence Equations~(\ref{twopoint}) and (\ref{fourpoint}) are trivially satisfied, with
\be\label{initial}\tag{R0}
\kert^{(1)}_{\alpha_1\alpha_2}=C_{b}^{(1)}+C_{W}^{(1)}\cdot\le(\frac{\inp_{\alpha_1}\cdot\inp_{\alpha_2}}{n_0}\ri)\, ,\ \ \ \SE_{\alpha_1\alpha_2}^{(1)}=0\, ,\ \ \ \text{and}\ \ \ \FPV^{(1)}_{(\alpha_1 \alpha_2) (\alpha_3 \alpha_4)}=0\, .
\ee
Obtained in Section~\ref{flow} are the recursive formulae that link these core kernel, self-energy, and four-point vertex at the $\ell$-th layer to those at the $(\ell+1)$-th layer while in Section~\ref{Bayes} these tensors at the last layer $\ell=L$ are used to yield the leading $1/n$ correction for Bayesian inference at finite widths.

\subsection{Related work}
Our Schwinger operator approach is orthogonal to the replica approach by~\cite{CMR2019} and, unlike the planar diagrammatic approach by~\cite{DG2019}, applies to general activation functions, made possible by accumulating corrections layer by layer rather than dealing with them all at once. See also \cite{Antognini2019}. More substantially, in contrast to these previous approaches, we here study finite-width effects on Bayesian inference and find that the renormalization-group picture naturally emerges, with layers playing the role of scales. 

\section{Distributional flow}\label{flow}
As auxiliary objects in recursive steps, let us introduce activation correlation functions
\be
\ACF^{(\ell)}_{i_1 \ldots i_m; \alpha_1 \ldots \alpha_m}\equiv \mathbb{E}\le\{\sigma\le[z_{i_1}^{(\ell)}\le(\inp_{\alpha_1}\ri)\ri]\cdots \sigma\le[z_{i_m}^{(\ell)}\le(\inp_{\alpha_m}\ri)\ri]\ri\}\, .
\ee
Our basic strategy is to establish relations
\be\label{philosophy}\tag{ZIGZAG}
\le\{\PCFt^{(1)}\ri\}\rightarrow\le\{\ACFt^{(1)}\ri\}\rightarrow \le\{\PCFt^{(2)}\ri\}\rightarrow\cdots\rightarrow \le\{\ACFt^{(L-1)}\ri\}\rightarrow \le\{\PCFt^{(L)}\ri\}\, ,
\ee
zigzagging between sets of preactivation correlation functions and sets of activation correlation functions, keeping track of leading finite-width corrections. Below, relations $\PCFt^{(\ell)}\rightarrow\ACFt^{(\ell)}$ are obtained by integrating out preactivations while relations $\ACFt^{(\ell)}\rightarrow\PCFt^{(\ell+1)}$ are obtained by integrating out biases and weights.
At first glance the algebra in this paper may look horrifying but repeated applications of Wick's contractions are all there is to it. The results are summarized in Section~\ref{master}.

\subsection{Zigzag relations for preactivation and activation correlation functions}\label{zigzag}
Integrating over the Gaussian biases and weights at $\ell$'s connections yield the relations that link activation correlations $\ACFt^{(\ell)}$ to preactivation correlations $\PCFt^{(\ell+1)}$ at the next layer. Recalling Equations~(\ref{twopoint}) and~(\ref{fourpoint}), trivial Wick's contractions yield
\bea\label{Htwo}
\kert_{\alpha_1 \alpha_2}^{(\ell+1)}+\frac{1}{n_{\ell}}\SE^{(\ell+1)}_{\alpha_1\alpha_2}&=&C_{b}^{(\ell+1)}+C_{W}^{(\ell+1)}\le[\frac{1}{n_{\ell}}\sum_{j=1}^{n_{\ell}}
\ACF^{(\ell)}_{j j; \alpha_1 \alpha_2}\ri]+O\le(\frac{1}{n^{2}}\ri)\, \ \ \ \text{and}\, \\
\FPV^{(\ell+1)}_{(\alpha_1 \alpha_2) (\alpha_3 \alpha_4)}&=&\frac{\le[C^{(\ell+1)}_{W}\ri]^2}{n_{\ell}}\sum_{j,k=1}^{n_{\ell}}\le[\ACF^{(\ell)}_{j j k k; \alpha_1 \alpha_2 \alpha_3 \alpha_4}-\ACF^{(\ell)}_{j j; \alpha_1 \alpha_2}\ACF^{(\ell)}_{k k; \alpha_3 \alpha_4}\ri]+O\le(\frac{1}{n}\ri)\, .\ \ \ \ \ \ \ \label{Hfour}
\eea
The remaining task is to relate preactivation correlations $\PCFt^{(\ell)}$ to activation correlations $\ACFt^{(\ell)}$ within the same layer, which will complete the zigzag relation~(\ref{philosophy}) for these correlation functions.\footnote{The nontrivial parts of the inductive proof for Equations~(\ref{twopoint}) and~(\ref{fourpoint}) are to show (i) that the right-hand side of Equation~(\ref{Hfour}) is finite as $n\rightarrow\infty$, (ii) that the leading contribution of Equation~(\ref{Htwo}) is the Gaussian-process kernel, and (iii) that higher-point connected preactivation correlation functions are all suppressed by $O\le(\frac{1}{n^{2}}\ri)$, all of which are verified in obtaining the recursive equations. See Appendix~\ref{activation_detail} for a full proof.}

With the mastery of Wick's contractions and connected correlation functions, it is simple to derive the following combinatorial hack (Appendix~\ref{hack}): viewing prior preactivations
\be
\rvz\equiv\le\{\rz_{i;\alpha}\equiv z^{(\ell)}_i\le(\inp_{\alpha}\ri)\ri\}_{i=1,\ldots,n_{\ell}; \alpha=1,\ldots,\ND}\, \nonumber
\ee
at the $\ell$-th layer as a random $\le(n_{\ell}\ND\ri)$-dimensional vector and defining the Gaussian integral with the kernel $\le\langle \rz_{i_1;\alpha_1}\rz_{i_2;\alpha_2}\ri\rangle_{\ker^{(\ell)}}=\ker^{(\ell)}_{i_1i_2; \alpha_1 \alpha_2}\equiv\delta_{i_1i_2}\kert_{\alpha_1 \alpha_2}^{(\ell)}$, the prior average
\be\label{hackeq}\tag{HACK}
\mathbb{E}\le\{\mathcal{F}[\rvz]\ri\}=\le\langle \mathcal{F}[\rvz]\ri\rangle_{\ker^{(\ell)}}+\frac{1}{n_{\ell-1}}\le[\le\langle \mathcal{F}[\rvz] \mathcal{O}_{S}[\rvz]+\mathcal{F}[\rvz] \mathcal{O}_{V}[\rvz]\ri\rangle_{\ker^{(\ell)}}\ri]+O\le(\frac{1}{n^2}\ri)
\ee
for any function $\mathcal{F}$. Here the operators $\mathcal{O}_{S}[\rvz]$ and $\mathcal{O}_{V}[\rvz]$ capture $1/n$ corrections due to self-energy and four-point vertex, respectively, and are defined as
\begin{align}
\mathcal{O}_{S}[\rvz]&\equiv \frac{1}{2}\sum_{\alpha_1,\alpha_2} \SE_{(\ell)}^{\alpha_1\alpha_2} \le[\le(\sum_{i=1}^{n_{\ell}} \rz_{i;\alpha_1}\rz_{i;\alpha_2}\ri)-n_{\ell}\kert^{(\ell)}_{\alpha_1\alpha_2}\ri]\, \ \ \ \text{and}\label{OS}\tag{OS}\\
\mathcal{O}_{V}[\rvz]&\equiv \frac{1}{8}\sum_{\alpha_1,\alpha_2,\alpha_3,\alpha_4} \FPV_{(\ell)}^{(\alpha_1\alpha_2)(\alpha_3\alpha_4)}\, \label{OV}\tag{OV}\\
&\times\Bigg[\le(\sum_{i=1}^{n_{\ell}} \rz_{i;\alpha_1}\rz_{i;\alpha_2}\ri)\le(\sum_{j=1}^{n_{\ell}} \rz_{j;\alpha_3}\rz_{j;\alpha_4}\ri)-2n_{\ell}\le(\sum_{i=1}^{n_{\ell}} \rz_{i;\alpha_1}\rz_{i;\alpha_2}\ri)\kert^{(\ell)}_{\alpha_3\alpha_4}\, \nonumber\\
&\ \ \ \ \ -4\le(\sum_{i=1}^{n_{\ell}} \rz_{i;\alpha_1}\rz_{i;\alpha_3}\ri)\kert^{(\ell)}_{\alpha_2 \alpha_4}+n_{\ell}^2\kert^{(\ell)}_{\alpha_1 \alpha_2}\kert^{(\ell)}_{\alpha_3 \alpha_4}+2n_{\ell}\kert^{(\ell)}_{\alpha_1 \alpha_3}\kert^{(\ell)}_{\alpha_2 \alpha_4}\Bigg]\, ,\nonumber
\end{align}
where the sample indices are raised by using the inverse core kernel as a metric, meaning
\bea
\SE_{(\ell)}^{\alpha_1 \alpha_2}&\equiv&\sum_{\alpha'_1,\alpha'_2}\le(\kert_{(\ell)}^{-1}\ri)^{\alpha_1 \alpha'_1}\le(\kert_{(\ell)}^{-1}\ri)^{\alpha_2 \alpha'_2}\SE^{(\ell)}_{\alpha'_1 \alpha'_2}\, \ \ \ \text{and}\,\\
\FPV_{(\ell)}^{(\alpha_1 \alpha_2) (\alpha_3 \alpha_4)}&\equiv&\sum_{\alpha'_1,\ldots,\alpha'_4}\le(\kert_{(\ell)}^{-1}\ri)^{\alpha_1 \alpha'_1}\cdots\le(\kert_{(\ell)}^{-1}\ri)^{\alpha_4 \alpha'_4}\FPV^{(\ell)}_{(\alpha'_1 \alpha'_2)(\alpha'_3 \alpha'_4)}\, .
\eea
Using the above hack, we can evaluate the activation correlations by straightforward algebra with Wick's contractions. In particular, as the Gaussian integral is diagonal in the neuron index $i$, we just need to disentangle cases with repeated and unrepeated neuron indices. The solution for this exercise is in Appendix~\ref{activation_detail}: it is arguably the most cumbersome algebra in this paper.

\subsection{Master recursive flow equations}\label{master}
Denoting the Gaussian integral with the core kernel $\le\langle \tilde{\rz}_{\alpha_1}\tilde{\rz}_{\alpha_2}\ri\rangle_{\kert^{(\ell)}}=\kert^{(\ell)}_{\alpha_1 \alpha_2}$ for a single-neuron random vector $\tilde{\rvz}\equiv\le\{\tilde{\rz}_{\alpha}\ri\}_{\alpha=1,\ldots,\ND}$, and plugging in results of Appendix~\ref{activation_detail} into Equations~(\ref{Htwo}) and (\ref{Hfour}), we arrive at our master recursion relations
\begin{align}
\kert_{\alpha_1 \alpha_2}^{(\ell+1)}&=C_{b}^{(\ell+1)}+C_{W}^{(\ell+1)}\le\langle \sigma(\tilde{\rz}_{\alpha_1}) \sigma(\tilde{\rz}_{\alpha_2})\ri\rangle_{\kert^{(\ell)}}\, ,\label{R1}\tag{R1}\\
\FPV^{(\ell+1)}_{(\alpha_1 \alpha_2) (\alpha_3 \alpha_4)}&=\le[C^{(\ell+1)}_{W}\ri]^2\Bigg[
\le\langle \sigma(\tilde{\rz}_{\alpha_1}) \sigma(\tilde{\rz}_{\alpha_2}) \sigma(\tilde{\rz}_{\alpha_3}) \sigma(\tilde{\rz}_{\alpha_4})\ri\rangle_{\kert^{(\ell)}}\, \label{R2}\tag{R2}\\
&\ \ \ \ \ \ \ \ \ \ \ \ \ \ \ \ \ \ \ \ \ \ \ -\le\langle \sigma(\tilde{\rz}_{\alpha_1}) \sigma(\tilde{\rz}_{\alpha_2})\ri\rangle_{\kert^{(\ell)}}\le\langle \sigma(\tilde{\rz}_{\alpha_3}) \sigma(\tilde{\rz}_{\alpha_4})\ri\rangle_{\kert^{(\ell)}}\,\nonumber\notag\\
&+\frac{1}{4}\le(\frac{n_{\ell}}{n_{\ell-1}}\ri)\sum_{\alpha'_1,\alpha'_2,\alpha'_3,\alpha'_4} \FPV_{(\ell)}^{(\alpha'_1\alpha'_2)(\alpha'_3\alpha'_4)}\le\langle \sigma(\tilde{\rz}_{\alpha_1}) \sigma(\tilde{\rz}_{\alpha_2})(\tilde{\rz}_{\alpha'_1}\tilde{\rz}_{\alpha'_2}-\kert^{(\ell)}_{\alpha'_1 \alpha'_2})\ri\rangle_{\kert^{(\ell)}}\, \nonumber\notag\\
&\ \ \ \ \ \ \ \ \ \ \ \ \ \ \ \ \ \ \ \ \ \ \ \ \ \ \ \ \ \ \ \ \ \ \ \ \ \ \ \ \ \ \ \ \ \ \ \ \times\le\langle \sigma(\tilde{\rz}_{\alpha_3}) \sigma(\tilde{\rz}_{\alpha_4})(\tilde{\rz}_{\alpha'_3}\tilde{\rz}_{\alpha'_4}-\kert^{(\ell)}_{\alpha'_3 \alpha'_4})\ri\rangle_{\kert^{(\ell)}}\Bigg]\, ,\ \ \ \text{and}\, \nonumber\notag\\
\SE_{\alpha_1 \alpha_2}^{(\ell+1)}&=\le(\frac{n_{\ell}}{n_{\ell-1}}\ri)C^{(\ell+1)}_{W}\Bigg[\frac{1}{2}\sum_{\alpha'_1,\alpha'_2} \SE_{(\ell)}^{\alpha'_1 \alpha'_2}\le\langle \sigma(\tilde{\rz}_{\alpha_1}) \sigma(\tilde{\rz}_{\alpha_2})(\tilde{\rz}_{\alpha'_1}\tilde{\rz}_{\alpha'_2}-\kert_{\alpha'_1 \alpha'_2}^{(\ell)})\ri\rangle_{\kert^{(\ell)}}\, \label{R3}\tag{R3}\\
&\ \ \ \ \ \ \ \ \ \ \ \ \ \ \ \ \ \ \ \ \ \ \ \ \ \ \ \ \ \ \ \ \ +\frac{1}{8}\sum_{\alpha'_1,\alpha'_2,\alpha'_3,\alpha'_4} \FPV_{(\ell)}^{(\alpha'_1 \alpha'_2) (\alpha'_3 \alpha'_4)}\Big\langle \sigma(\tilde{\rz}_{\alpha_1}) \sigma(\tilde{\rz}_{\alpha_2})\,\nonumber\\
&\ \ \ \ \ \ \ \ \ \ \ \ \ \ \ \ \ \ \ \ \ \ \ \ \ \ \ \ \ \ \ \ \ \ \ \times\Big(\tilde{\rz}_{\alpha'_1}\tilde{\rz}_{\alpha'_2}\tilde{\rz}_{\alpha'_3}\tilde{\rz}_{\alpha'_4}-2\tilde{\rz}_{\alpha'_1}\tilde{\rz}_{\alpha'_2}\kert_{\alpha'_3 \alpha'_4}^{(\ell)}-4\tilde{\rz}_{\alpha'_1}\tilde{\rz}_{\alpha'_3}\kert_{\alpha'_2 \alpha'_4}^{(\ell)}\, \nonumber\\
&\ \ \ \ \ \ \ \ \ \ \ \ \ \ \ \ \ \ \ \ \ \ \ \ \ \ \ \ \ \ \ \ \ \ \ \ \ \ \ \ \ +\kert_{\alpha'_1\alpha'_2}^{(\ell)}\kert_{\alpha'_3\alpha'_4}^{(\ell)}+2\kert_{\alpha'_1\alpha'_3}^{(\ell)}\kert_{\alpha'_2\alpha'_4}^{(\ell)}\Big)\Big\rangle_{\kert^{(\ell)}}\Bigg]\, .\nonumber
\end{align}
For $\ell=1$, a special note about the ratio $\frac{n_{\ell}}{n_{\ell-1}}$ is in order: even though $n_{0}$ stays constant while $n_{1}\gg1$, the terms proportional to that ratio are identically zero due to the complete Gaussianity~(\ref{initial}).

The preactivation distribution in the first layer~(\ref{initial}) sets the initial condition for the flow from lower to higher layers dictated by these recursive equations. Evolving through these recursive equations, the running couplings -- $\kert_{\alpha_1 \alpha_2}^{(\ell)}$, $\FPV^{(\ell)}_{(\alpha_1 \alpha_2) (\alpha_3 \alpha_4)}$, and $\SE_{\alpha_1 \alpha_2}^{(\ell)}$ -- then trace changes in the distributions of preactivations as the layer scale $\ell$ shifts, just as running couplings for physical systems track changes in effective Boltzmann distributions as the probing scale shifts.
Once recursed up to the last layer $\ell=L$, the resulting distribution of outputs $\rvz=\rvz^{(L)}$ can be succinctly encoded by the probability distribution
\be\label{D0}\tag{D0}
p[\rvz]=\frac{e^{-\ac[\rvz]}}{\int \mathrm{d}\rvz'e^{-\ac[\rvz']} }\, 
\ee
with the potential $\ac[\rvz]=\ac_{0}[\rvz]+\epsilon\ac_1[\rvz]+O(\epsilon^2)$ where $\epsilon\equiv\frac{1}{n_{L-1}}\ll1$,
\begin{align}
\ac_0[\rvz]=&\frac{1}{2}\sum_{\alpha_1,\alpha_2} \le(\kert_{(L)}^{-1}\ri)^{\alpha_1 \alpha_2}\le(\sum_{i=1}^{n_{L}} \rz_{i;\alpha_1}\rz_{i;\alpha_2}\ri)\, ,  \ \ \text{and}\, \label{D1}\tag{D1}\\\
\ac_1[\rvz]=&-\frac{1}{2}\sum_{\alpha_1,\alpha_2} \widetilde{J}^{\alpha_1 \alpha_2}\le(\sum_{i=1}^{n_{L}} \rz_{i;\alpha_1}\rz_{i;\alpha_2}\ri)\, \label{D2}\tag{D2}\\
&-\frac{1}{8}\sum_{\alpha_1,\alpha_2,\alpha_3,\alpha_4} \FPV_{(L)}^{(\alpha_1 \alpha_2) (\alpha_3 \alpha_4)}\le(\sum_{i=1}^{n_{L}} \rz_{i;\alpha_1}\rz_{i;\alpha_2}\ri)\le(\sum_{j=1}^{n_{L}} \rz_{j;\alpha_3}\rz_{j;\alpha_4}\ri)\, \ \ \ \text{with}\, \nonumber\\
\widetilde{J}^{\alpha_1 \alpha_2}\equiv& \SE_{(L)}^{\alpha_1 \alpha_2}-\sum_{\alpha_3,\alpha_4}\kert^{(L)}_{\alpha_3 \alpha_4}\le[\FPV_{(L)}^{(\alpha_1 \alpha_3) (\alpha_2 \alpha_4)}+\frac{n_L}{2}\FPV_{(L)}^{(\alpha_1 \alpha_2) (\alpha_3 \alpha_4)}\ri]\, .
\end{align}
Again, this can be derived through Wick's contractions. It is important to note that $n_L$ is constant and thus $\epsilon\ac_{1}[\rvz]$ can consistently be treated perturbatively.\footnote{If $n_{L}$ were of order $n\gg1$, the potential $\ac$ would become a large-$n$ vector model, for which we would have to sum the infinite series of bubble diagrams~\cite{MZ2003}.
}

\section{Interlude: examples}\label{examples}
The recursive relations obtained above can be evaluated numerically~\cite{LBNSPS2017} [or sometimes analytically for rectified linear unit (ReLU) activation~\cite{CS2009}], which is a perfectly adequate approach: at the leading order it involves four-dimensional Gaussian integrals at most. Here, continuing the theme of wearing out Wick's contractions, we develop an alternative analytic method that works for any polynomial activations~\cite{LP2017}, providing another perfectly cromulent approach.

For a general polynomial activation of degree $p$, $\sigma(z)=\sum_{k=0}^{p} a_k z^k$, the nontrivial term in Equation~(\ref{R1}) can be expanded as
\be
\le\langle \sigma(\tilde{\rz}_{\alpha_1}) \sigma(\tilde{\rz}_{\alpha_2})\ri\rangle_{\kert^{(\ell)}}=\sum_{k_1,k_2=0}^{p} a_{k_1} a_{k_2}\le\langle \le(\tilde{\rz}_{\alpha_1}\ri)^{k_1} \le(\tilde{\rz}_{\alpha_2}\ri)^{k_2}\ri\rangle_{\kert^{(\ell)}}\, .
\ee
Each term can then be evaluated by Wick's contractions and the same goes for all the terms in Equations~(\ref{R2}) and~(\ref{R3}).\footnote{The same approach could be adopted for an analytic function but it would in general be difficult to sum the resulting infinite series in a closed form. It could nonetheless be useful in, for example, proving convergence properties.} Below and in Appendix~\ref{bestiary}, we illustrate this procedure with simple examples.

\subsection{Deep linear networks}\label{linear}
When the activation function is linear, $\sigma(z)=z$, multilayer perceptrons are called deep linear networks~\cite{SMG2013}. Setting $C_{b}^{(\ell)}=0$ and $C_{W}^{(\ell)}=1$ for simplicity, our recursion relations reduce to $\kert_{\alpha_1 \alpha_2}^{(\ell+1)}=\kert_{\alpha_1 \alpha_2}^{(\ell)}$, 
\be
\FPV^{(\ell+1)}_{(\alpha_1 \alpha_2) (\alpha_3 \alpha_4)}=\le[\kert_{\alpha_1 \alpha_3}^{(\ell)}\kert_{\alpha_2 \alpha_4}^{(\ell)}+\kert_{\alpha_1 \alpha_4}^{(\ell)}\kert_{\alpha_2 \alpha_3}^{(\ell)}+\le(\frac{n_{\ell}}{n_{\ell-1}}\ri)\FPV^{(\ell)}_{(\alpha_1 \alpha_2) (\alpha_3 \alpha_4)}\ri]\, ,\nonumber
\ee
and $\SE^{(\ell+1)}_{\alpha_1 \alpha_2}=\le(\frac{n_{\ell}}{n_{\ell-1}}\ri)\SE^{(\ell)}_{\alpha_1\alpha_2}$. Solving them yields the layer-independent core kernel and zero self-energy
\be\label{KSElinear}
\kert_{\alpha_1 \alpha_2}^{(\ell)}=\kert_{\alpha_1 \alpha_2}^{(1)}=\frac{\inp_{\alpha_1}\cdot\inp_{\alpha_2}}{n_0}\, \ \ \ \text{and}\ \ \ \SE_{\alpha_1 \alpha_2}^{(\ell)}=0
\ee
and the linearly layer-dependent four-point vertex
\be
\frac{1}{n_{\ell-1}}\FPV^{(\ell)}_{(\alpha_1 \alpha_2) (\alpha_3 \alpha_4)}=\le(\sum_{\ell'=1}^{\ell-1}\frac{1}{n_{\ell'}}\ri)\le[\kert_{\alpha_1 \alpha_3}^{(1)}\kert_{\alpha_2 \alpha_4}^{(1)}+\kert_{\alpha_1 \alpha_4}^{(1)}\kert_{\alpha_2 \alpha_3}^{(1)}\ri]\, .
\ee
It succinctly reproduces the result that can be obtained through planar diagrams in this special setup~\cite{DG2019}. Quadratic activation~\cite{LMZ2017} is worked out in Appendix~\ref{bestiary_quad}. 

\subsection{ReLU with single input}\label{relu_single}
The recursion relations simplify drastically for the case of a single input, $\ND=1$, as worked out in detail in Appendix~\ref{bestiary_mono}. For instance, for ReLU activation with $C_{b}^{(\ell)}=0$ and $C_{W}^{(\ell)}=2$, we obtain the layer-independent core kernel, zero self-energy, and the four-point vertex
\be
\frac{1}{n_{\ell-1}}\FPV^{(\ell)}_{(\alpha \alpha) (\alpha \alpha)}=5\le(\sum_{\ell'=1}^{\ell-1}\frac{1}{n_{\ell'}}\ri)\le(\kert_{\alpha \alpha}^{(1)}\ri)^2\, .
\ee
Interestingly, as for deep linear networks, the factor $\sum_{\ell'}(1/n_{\ell'})$ appears again. This factor has also been found by~\cite{HR2018}, which provides guidance for network architectural design through its minimization. We generalize this factor for monomial activations in Appendix~\ref{monomial_single}

\subsection{Experimental verification: output distributions for a single input}\label{Fexperiment}
\begin{figure}[h]
\centering{
 \includegraphics[width=0.5\linewidth]{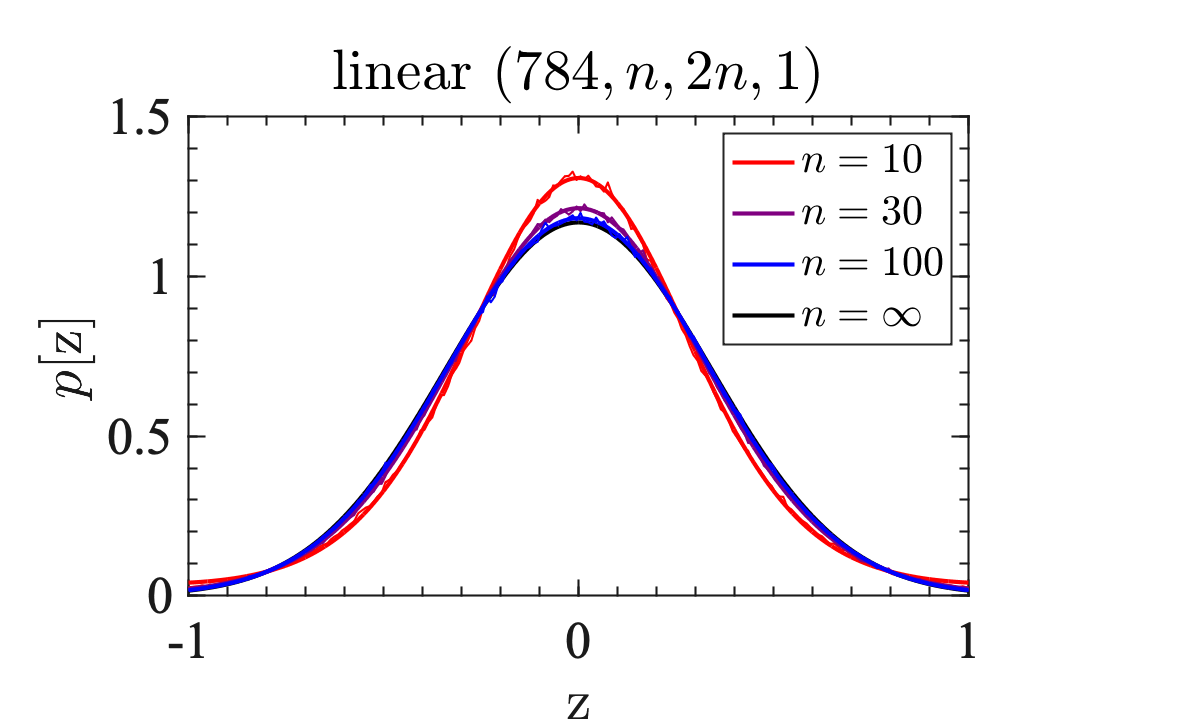}\hspace{-0.8cm}
 \includegraphics[width=0.5\linewidth]{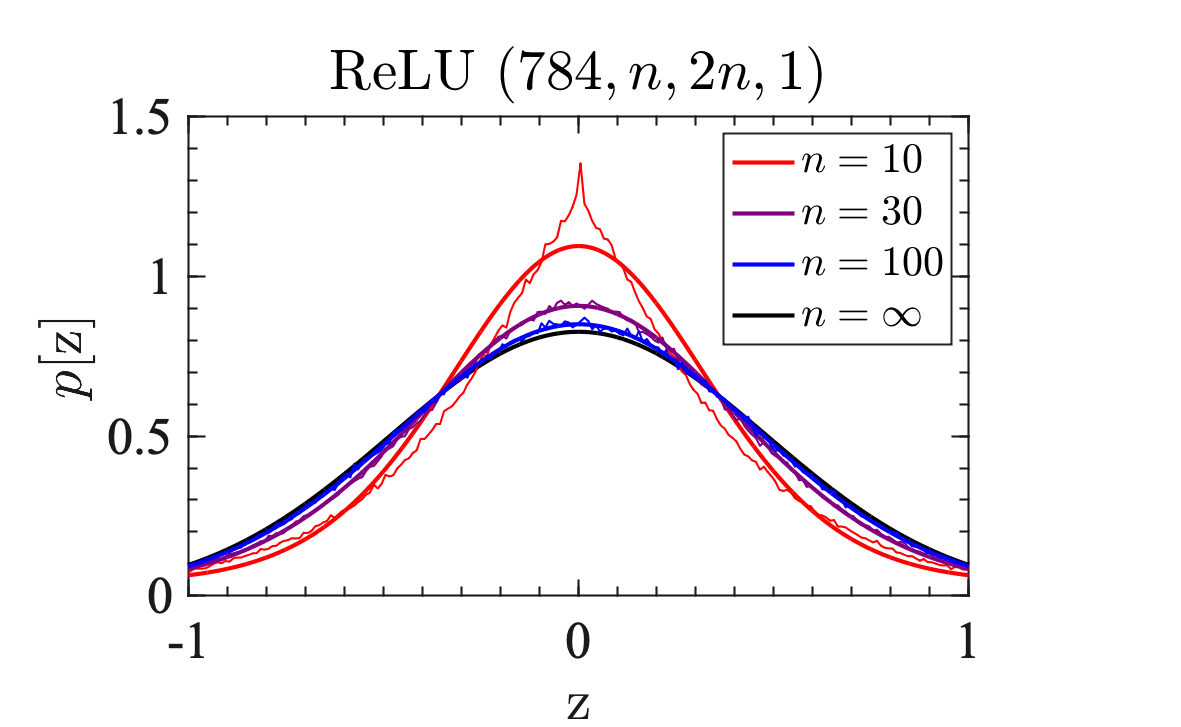}
}
\caption{Comparison between theory and experiments for prior distributions of outputs for a single input. The agreement between our theoretical predictions (smooth thick lines) and experimental data (rugged thin lines) is superb, correctly capturing the initial deviations from Gaussian processes at $n=\infty$ (black), all the way down to $n\sim10$ for linear activation and to $n\sim 30$ for ReLU activation.}
\label{NNNGP1}
\end{figure}

Here we put our theory to the test. For concreteness, take a single black-white image of hand-written digits with $28$-by-$28$ pixels (i.e.~$n_0=784$) from the MNIST dataset~\cite{LBBH1998} without preprocessing, set depth $L=3$, bias variance $C^{(\ell)}_{b}=0$, weight variance $C^{(\ell)}_W=C_W$, and widths $(n_0,n_1,n_2,n_3)=(784,n,2n,1)$, and use activations $\sigma(z)=z$ (linear) with $C_W=1$ and $\mathrm{max}(0,z)$ (ReLU) with $C_W=2$. In Figure~\ref{NNNGP1}, for each width-parameter $n$ of the hidden layers we record the prior distribution of outputs over $10^6$ instances of Gaussian weights and compare it with the theoretical prediction -- obtained by cranking the knob from the initial condition~(\ref{initial}) through the recursion relations~(\ref{R1}-\ref{R3}) to the distribution~(\ref{D0}-\ref{D2}). The prior distribution becomes increasingly non-Gaussian as networks narrow and the deviation from the Gaussian-process prior is correctly captured by our theory. Higher-order perturbative calculations are expected to systematically improve the quality -- and extend the range -- of the agreement. Additional experiments are performed in Appendix~\ref{more_experiments}, which further corroborates our theory.


\section{Bayesian inference}\label{Bayes}
Let us take off from the terminal point of Section~\ref{flow}: we have obtained the recursive equations~(\ref{initial}-\ref{R3}) for the Gaussian-process kernel and the leading finite-width corrections and codified them in the weakly non-Gaussian prior distributions $p[\rvz]$~(\ref{D0}-\ref{D2}) of outputs
\be
\rvz\equiv\le\{\rz_{i;\alpha}\equiv z^{(L)}_i\le(\inp_{\alpha}\ri)\ri\}_{i=1,\ldots,n_{L}; \alpha=1,\ldots,\ND}\, ,\nonumber
\ee
dictated by the potential $\ac[\rvz]=\ac_{0}[\rvz]+\epsilon\ac_1[\rvz]+O(\epsilon^2)$ with $\epsilon\equiv\frac{1}{n_{L-1}}\ll1$. Examples in Section~\ref{examples} illustrate that finite-width corrections stay perturbative typically when $\frac{\text{depth}}{\text{width}}\ll 1$.
Let us now divide $\ND$ inputs into $\NR$ training and $\NE$ test inputs as\be
\le\{\inp_{\alpha}\ri\}_{\alpha=1,\ldots,\ND}=\{\le(\inp_{\text{R}}\ri)_{\tra}\}_{\tra= 1,\ldots,\NR}\cup\{\le(\inp_{\text{E}}\ri)_{\tes}\}_{\tes=1,\ldots,\NE}\, ,
\ee
and the training inputs come with target outputs
\be
\{\le(\YC\ri)_{\tra}\}_{\tra=1,\ldots,\NR}=\{\le(\tar_{\text{R}}\ri)_{i;\tra}\}_{\tra=1,\ldots,\NR; i=1,\ldots,n_{L}}\, .
\ee
We shall develop a procedure to infer outputs for test inputs a l\'{a} Bayes, perturbatively extending the textbook~\cite{WR2006}. For field theorists, our calculation is just a background-field calculation~\cite{Weinberg1996} in disguise.

Taking the liberty of notations, we let the number of input-data arguments dictate the summation over sample indices $\alpha$ inside the potential $\ac$, and denote the joint probabilities
\be
p[\rvz_{\text{R}}]=\frac{e^{-\ac[\rvz_{\text{R}}]}}{\int \mathrm{d} \rvz'_{\text{R}} e^{-\ac[\rvz'_{\text{R}}]}}\ \ \ \text{and}\ \ \ p[\rvz_{\text{R}},\rvz_{\text{E}}]=\frac{e^{-\ac[\rvz_{\text{R}},\rvz_{\text{E}}]}}{\int \mathrm{d} \rvz'_{\text{R}}\mathrm{d} \rvz'_{\text{E}} e^{-\ac[\rvz'_{\text{R}},\rvz'_{\text{E}}]}}\, .
\ee
Given the training targets $\YC$, the posterior distribution of test outputs are given by Bayes' rule:
\be\label{bayesP}\tag{Bayes}
p\le[\rvz_{\text{E}}\vert \YC\ri]=\frac{p[\YC,\rvz_{\text{E}}]}{p[\YC]}\, =\le(\frac{\int \mathrm{d} \rvz'_{\text{R}} e^{-\ac[\rvz'_{\text{R}}]}}{\int \mathrm{d} \rvz'_{\text{R}}\mathrm{d} \rvz'_{\text{E}} e^{-\ac[\rvz'_{\text{R}},\rvz'_{\text{E}}]}}\ri)e^{-\le(\ac[\YC,\rvz_{\text{E}}]-\ac[\YC]\ri)}\, .
\ee
The leading Gaussian-process contributions can be segregated out through the textbook manipulation~\cite{WR2006} [c.f.~Appendix~\ref{NNNGPmanipulation}]: denoting the full Gaussian-process kernel in the last layer as
\be
\ker_{i_1 i_2; \alpha_1 \alpha_2}=\delta_{i_1 i_2}
\begin{pmatrix}
\le(\kert_{\text{RR}}\ri)_{\tra_1 \tra_2} & \le(\kert_{\text{RE}}\ri)_{\tra_1 \tes_2}\\
\le(\kert_{\text{ER}}\ri)_{\tes_1 \tra_2} & \le(\kert_{\text{EE}}\ri)_{\tes_1 \tes_2}
\end{pmatrix}\, 
\ee
and the Gaussian-process posterior mean prediction as
\be\label{GPP}\tag{GPM}
\le(\GPM\ri)_{i;\tes}\equiv\sum_{\tra}\le[\kert_{\text{ER}} \kert_{\text{RR}}^{-1}\ri]_{\tes}^{\ \ \tra} \le(\tar_{\text{R}}\ri)_{i;\tra} \, ,
\ee
and defining a fluctuation $\le(\rz_{\text{E}}\ri)_{i;\tes}\equiv\le(\GPM\ri)_{i;\tes}+\le(\delta \rz_{\text{E}}\ri)_{i;\tes}$ and a matrix $\kert_{\Delta}\equiv \kert_{\text{EE}}- \kert_{\text{ER}} \kert_{\text{RR}}^{-1} \kert_{\text{RE}}$,
\be\label{GPH}\tag{GP$\Delta$}
\ac_{0}\le[\YC,\rvz_{\text{E}}\ri]-\ac_{0}\le[\YC\ri]=\frac{1}{2}\sum_{i}\sum_{\tes_1,\tes_2}\le(\delta \rz_{\text{E}}\ri)_{i;\tes_1}\le(\kert_{\Delta}^{-1}\ri)^{\tes_1\tes_2}\le(\delta \rz_{\text{E}}\ri)_{i;\tes_2}\, .
\ee
For any function $\mathcal{F}$, its expectation over the Bayesian posterior~(\ref{bayesP}) then turns into
\be
\int \mathrm{d}\rvz_{\text{E}} \mathcal{F}[\rvz_{\text{E}}] p\le[\rvz_{\text{E}}\vert \YC\ri]=\widetilde{\mathcal{N}}\le\langle e^{-\epsilon \ac_{1}\le[\YC,\mathbf{y}_{\text{E}}^{\text{GP}}+\delta \rvz_{\text{E}}\ri]} \mathcal{F}[\mathbf{y}_{\text{E}}^{\text{GP}}+\delta \rvz_{\text{E}}]\ri\rangle_{\ker_{\Delta}}
\ee
where the deviation kernel $\le\langle \le(\delta \rz_{\text{E}}\ri)_{i_1;\tes_1} \le(\delta \rz_{\text{E}}\ri)_{i_2;\tes_2}\ri\rangle_{\ker_{\Delta}}\equiv\delta_{i_1 i_2}\le(\kert_{\Delta}\ri)_{\tes_1 \tes_2}$ and the normalization factor
\be
\widetilde{\mathcal{N}}=\le[\le\langle e^{-\epsilon \ac_{1}\le[\YC,\mathbf{y}_{\text{E}}^{\text{GP}}+\delta \rvz_{\text{E}}\ri]}\ri\rangle_{\ker_{\Delta}}\ri]^{-1}=1+O(\epsilon)\, .
\ee
In particular the mean posterior output is given by
\bea
\le(\YT\ri)_{i;\tes}\equiv\int \mathrm{d}\rvz_{\text{E}} \le(\rz_{\text{E}}\ri)_{i;\tes} p\le[\rvz_{\text{E}}\vert \YC\ri]&=&\le(\GPM\ri)_{i;\tes}+\widetilde{\mathcal{N}}\le\langle \le(\delta \rz_{\text{E}}\ri)_{i;\tes} e^{-\epsilon \ac_{1}\le[\YC, \mathbf{y}_{\text{E}}^{\text{GP}}+\delta \rvz_{\text{E}}\ri]}\ri\rangle_{\ker_{\Delta}}\, \\
&=&\le(\GPM\ri)_{i;\tes}-\epsilon\le\langle  \le(\delta\rz_{\text{E}}\ri)_{i;\tes}  \ac_{1}\le[\YC,\mathbf{y}_{\text{E}}^{\text{GP}}+\delta \rvz_{\text{E}}\ri]\ri\rangle_{\ker_{\Delta}}+O(\epsilon^2)\, . \nonumber
\eea
Stringing together $\overline{\phi}_{i;\alpha}\equiv [\le(\tar_{\text{R}}\ri)_{i;\tra},\le(\GPM\ri)_{i;\tes}]$, recalling Equation~(\ref{D2}) for $\mathcal{H}_1$, and using Wick's contractions for one last time, the mean prediction becomes
\begin{align}
&\le(\GPM\ri)_{i;\tes}\, \label{NNNGPP}\tag{NGPM}\\
&+\epsilon \sum_{\alpha_1,\tes_1}\le(\kert_{\Delta}\ri)_{\tes \tes_1}\overline{\phi}_{i;\alpha_1}\Bigg[\SE^{\tes_1\alpha_1}-\sum_{\alpha_2,\alpha_3}\FPV^{(\tes_1\alpha_2)(\alpha_1\alpha_3)}\kert_{\alpha_2\alpha_3}+\sum_{\tes_2,\tes_3}\FPV^{(\tes_1\tes_2)(\alpha_1\tes_3)}\le(\kert_{\Delta}\ri)_{\tes_2 \tes_3}\, \nonumber\\
&+\frac{n_{L}}{2}\sum_{\tes_2,\tes_3}\FPV^{(\alpha_1\tes_1)(\tes_2\tes_3)}\le(\kert_{\Delta}\ri)_{\tes_2 \tes_3}+\sum_{\alpha_2,\alpha_3}\FPV^{(\tes_1 \alpha_1) (\alpha_2 \alpha_3)}\le(-\frac{n_L}{2}\kert_{\alpha_2 \alpha_3}+\frac{1}{2}\sum_{j}\overline{\phi}_{j;\alpha_2}\overline{\phi}_{j;\alpha_3}\ri)\Bigg]\, .\nonumber
\end{align}



\begin{figure}[h]
\centering{
 \includegraphics[width=0.35\linewidth]{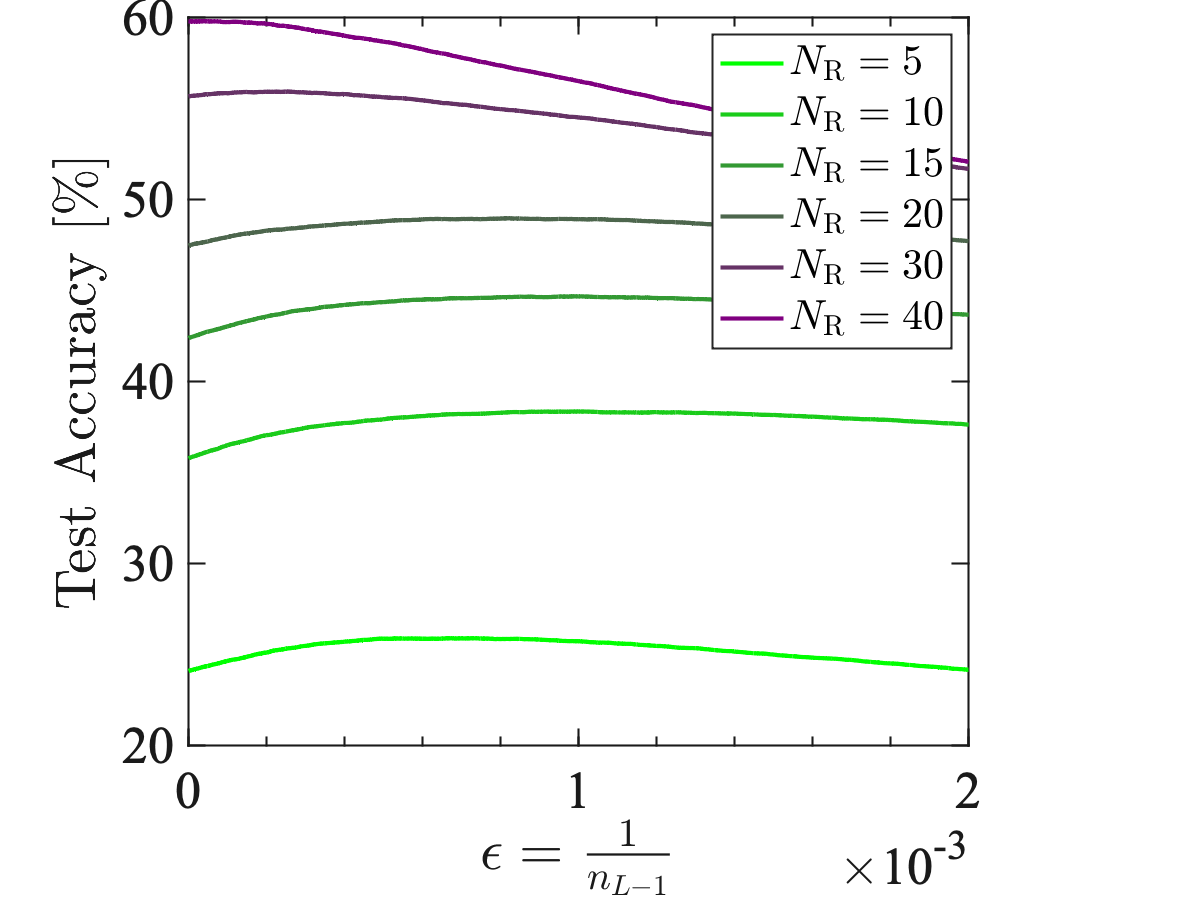}\hspace{-0.8cm}
}
\caption{Test accuracy for $\NE=10000$ MNIST test data as a function of the  inverse width $\epsilon=1/n_{L-1}$ of the hidden layer with quadratic activation. For each number $\NR$ of subsampled training data, the result is averaged over $10$ distinct choices of such subsamplings. For small numbers of training data, finite widths result in regularization effects, improving the test accuracy.}
\label{NNNGP2}
\end{figure}

With additional manipulations, this expression can be simplified into the actionable form that is amenable to use in practice [c.f.~Equations~(\ref{NNNGPPdash}) and~(\ref{NNNGPPdashdash}) in Appendix~\ref{NNNGPmanipulation}].
It turns out that for deep linear networks the leading finite-width correction vanishes, and the first correction is likely to show up at higher order in $1/n$ asymptotic expansion, which is not carried out in this paper. Here we instead use the $L=2$ multilayer perceptron with the quadratic activation $\sigma(z)=z^2$, zero bias variance $C^{(\ell)}_b=0$, and weight variance $C^{(\ell)}_W=1/3$ for illustration, plugging Equations~(\ref{R1quad},\ref{R2quad},\ref{R3quad}) into Equations~(\ref{NNNGPPdash}) and~(\ref{NNNGPPdashdash}) and varying $\epsilon\equiv \frac{1}{n_{L-1}}=\frac{1}{n_1}$. Results in Figure~\ref{NNNGP2} indicate the regularization effects of finite widths when the number of training samples, $\NR$, is small, resulting in peak performance at finite widths. This is in line with expectations that finite widths ameliorate overfitting and that non-Gaussian priors increase the expressivity of neural functions, but additional large-scale extensive experiments would be desirable in the future.

\section{Dreams}\label{conclusion}
In this paper, we have developed the perturbative formalism that captures the flow of preactivation distributions from lower to higher layers. The resemblance between our recursive equations and renormalization-group flow equations in high-energy and statistical physics is highly appealing. It would be exciting to investigate the structure of fixed points away from the Gaussian asymptopia~\cite{SGGS2016} and fully realize the dream articulated in Ref.~\cite{MS2014} -- the audacious hypothesis that neural networks wash away microscopic irrelevancies and extract relevant features -- beyond their limited example of a mapping between two antiquated techniques.

In addition we have developed the perturbative Bayesian inference scheme universally applicable whenever prior distributions are weakly non-Gaussian, and have applied it to the specific cases of neural networks at finite widths. In light of possible finite-width regularization effects, it would be prudent to revisit the empirical comparison between SGD optimization and Bayesian inference at finite widths~\cite{LBNSPS2017,NXBLYAPS2019}, especially for convolutional neural networks.

Finally, given surging interests in SGD dynamics within the large-width regime~\cite{JGH2018,COB2019,LXSBSP2019,CMR2019,DG2019}, it would be natural to adapt our formalism for investigating corrections to neural tangent kernels, and even aspire to capture a transition from lazy-learning to feature-learning regimes.
\section*{Acknowledgments}
The author thanks Yasaman Bahri for the discussion that seeded the idea for this project, Boris L.~Hanin for persistently preaching about Gaussian processes, and David J.~Schwab for permission to call his example limited with our friendship intact. The author also thanks Ethan S.~Dyer, Mario Geiger, Guy Gur-Ari, Eric T.~Mintun, Stephen H.~Shenker, and Lexing Ying for substantially useful discussions, and Daniel A.~Roberts for the quality control of all the jokes and more.

\bibliography{NNNGP}

\newpage
\appendix
\renewcommand\thefigure{S\arabic{figure}}
\setcounter{figure}{0}
\renewcommand\theequation{S\arabic{equation}}
\setcounter{equation}{0}

\section{Wick's tricks}\label{cumulants_for_dummies}
Here is all you need to know in order to follow the calculations in the paper.  In the main text, Wick's contractions are used both for trivially integrating out biases and weights as straightforward applications of Appendix~\ref{Wick} and for nontrivially integrating out preactivations, with concepts of cumulants reviewed in Appendix~\ref{CC} and~\ref{HC}, culminating in the hack derived in Appendix~\ref{hack}. The random variables are generically indexed by $\mu=1,\ldots,N$ throughout this Appendix: when applying formulae for biases, $\mu=i$; for weights $\mu=(i,j)$; for full preactivations $\mu=(i,\alpha)$; for single-neuron preactivations $\mu=\alpha$.

\subsection{Wick's contractions}\label{Wick}
For Gaussian-distributed variables $\rvz=\le\{\rz_{\mu}\ri\}_{\mu=1,\ldots,N}$ with a kernel $\ker_{\mu \mu'}$, moments
\be\label{GI}
\le\langle \rz_{\mu_1}\rz_{\mu_2} \cdots \rz_{\mu_m}\ri\rangle_{\ker}\equiv\frac{\int \mathrm{d}\rvz e^{-\ac_0[\rvz]}\rz_{\mu_1}\rz_{\mu_2}\cdots \rz_{\mu_m}}{\int \mathrm{d}\rvz e^{-\ac_0[\rvz]}}\, \ \ \ \text{with}\ \ \ \ac_0[\rvz]\equiv\frac{1}{2}\sum_{\mu,\mu'=1}^N \rz_{\mu} \le(\ker^{-1}\ri)^{\mu \mu'}\rz_{\mu'} \, .
\ee
For any odd $m$ such moments identically vanish. For even $m$, Isserlis-Wick's theorem states that
\be
\le\langle \rz_{\mu_1}\rz_{\mu_2} \cdots \rz_{\mu_m}\ri\rangle_{\ker}= \sum_{\text{all\ pairing}} \ker_{\mu_{k_1} \mu_{k_2}}\cdots \ker_{\mu_{k_{m-1}} \mu_{k_m}}
\ee
where the sum is over all the possible pairings of $m$ variables, $(k_1,k_2),\ldots,(k_{m-1},k_m)$. In general, there are $(m-1)!!=(m-1)\cdot(m-3)\cdots  1$ such pairings. For a proof, see for example~\cite{Zee2010}. In order to understand and use the theorem, it is instructive to look at a few examples:
\bea
\le\langle \rz_{\mu_1}\rz_{\mu_2}\ri\rangle_{\ker}&=&\ker_{\mu_1 \mu_2}\, ;\\
\le\langle \rz_{\mu_1}\rz_{\mu_2}\rz_{\mu_3}\rz_{\mu_4}\ri\rangle_{\ker}&=&\ker_{\mu_1 \mu_2}\ker_{\mu_3 \mu_4}+\ker_{\mu_1 \mu_3}\ker_{\mu_2 \mu_4}+\ker_{\mu_1 \mu_4}\ker_{\mu_2 \mu_3}\, ;
\eea
and
\bea
&&\le\langle \rz_{\mu_1}\rz_{\mu_2}\rz_{\mu_3}\rz_{\mu_4}\rz_{\mu_5}\rz_{\mu_6}\ri\rangle_{\ker}\, \\
&=&\ker_{\mu_1 \mu_2}\ker_{\mu_3 \mu_4}\ker_{\mu_5 \mu_6}+\ker_{\mu_1 \mu_3}\ker_{\mu_2 \mu_4}\ker_{\mu_5 \mu_6}+\ker_{\mu_1 \mu_4}\ker_{\mu_2 \mu_3}\ker_{\mu_5 \mu_6}\, \nonumber\\
&+&\ker_{\mu_1 \mu_2}\ker_{\mu_3 \mu_5}\ker_{\mu_4 \mu_6}+\ker_{\mu_1 \mu_3}\ker_{\mu_2 \mu_5}\ker_{\mu_4 \mu_6}+\ker_{\mu_1 \mu_5}\ker_{\mu_2 \mu_3}\ker_{\mu_4 \mu_6}\, \nonumber\\
&+&\ker_{\mu_1 \mu_2}\ker_{\mu_5 \mu_4}\ker_{\mu_3 \mu_6}+\ker_{\mu_1 \mu_5}\ker_{\mu_2 \mu_4}\ker_{\mu_3 \mu_6}+\ker_{\mu_1 \mu_4}\ker_{\mu_2 \mu_5}\ker_{\mu_3 \mu_6}\, \nonumber\\
&+&\ker_{\mu_1 \mu_5}\ker_{\mu_3 \mu_4}\ker_{\mu_2 \mu_6}+\ker_{\mu_1 \mu_3}\ker_{\mu_5 \mu_4}\ker_{\mu_2 \mu_6}+\ker_{\mu_1 \mu_4}\ker_{\mu_5 \mu_3}\ker_{\mu_2 \mu_6}\, \nonumber\\
&+&\ker_{\mu_5 \mu_2}\ker_{\mu_3 \mu_4}\ker_{\mu_1 \mu_6}+\ker_{\mu_5 \mu_3}\ker_{\mu_2 \mu_4}\ker_{\mu_1 \mu_6}+\ker_{\mu_5 \mu_4}\ker_{\mu_2 \mu_3}\ker_{\mu_1 \mu_6}\, .\nonumber
\eea

\subsection{Connected correlations}\label{CC}
Given general (not necessarily Gaussian) random variables, connected correlation functions are defined inductively through
\bea\label{cumu}
&&\mathbb{E}\le[\rz_{\mu_1} \rz_{\mu_2}\cdots \rz_{\mu_m}\ri]\, \\
&\equiv&\mathbb{E}\le[\rz_{\mu_1} \rz_{\mu_2}\cdots \rz_{\mu_m}\ri]\big|_{\text{connected}}\, \nonumber\\
&&+\sum_{\text{all\ subdivisions}}\mathbb{E}\le[\rz_{\mu_{k^{[1]}_1}}\cdots \rz_{\mu_{k^{[1]}_{\nu_1}}}\ri]\Big|_{\text{connected}}\cdots\mathbb{E}\le[\rz_{\mu_{k^{[s]}_1}} \cdots \rz_{\mu_{k^{[s]}_{\nu_s}}}\ri]\Big|_{\text{connected}}\, \nonumber
\eea
where the sum is over all the possible subdivisions of $m$ variables into $s>1$ clusters of sizes $(\nu_1,\ldots,\nu_s)$ as $(k^{[1]}_1,\ldots,k^{[1]}_{\nu_1}),\ldots,(k^{[s]}_1,\ldots,k^{[s]}_{\nu_s})$. In order to understand the definition, it is again instructive to look at a few examples. Assuming that all the odd moments vanish,
\bea
\mathbb{E}\le[\rz_{\mu_1} \rz_{\mu_2}\ri]&=&\mathbb{E}\le[\rz_{\mu_1} \rz_{\mu_2}\ri]\big|_{\text{connected}}\, \ \ \ \text{and}\, \\
\mathbb{E}\le[\rz_{\mu_1} \rz_{\mu_2}\rz_{\mu_3}\rz_{\mu_4}\ri]&=&\mathbb{E}\le[\rz_{\mu_1} \rz_{\mu_2}\rz_{\mu_3}\rz_{\mu_4}\ri]\big|_{\text{connected}}\, \\
&&+\mathbb{E}\le[\rz_{\mu_1} \rz_{\mu_2}\ri]\big|_{\text{connected}}\mathbb{E}\le[\rz_{\mu_3} \rz_{\mu_4}\ri]\big|_{\text{connected}}\, \nonumber\\
&&+\mathbb{E}\le[\rz_{\mu_1} \rz_{\mu_3}\ri]\big|_{\text{connected}}\mathbb{E}\le[\rz_{\mu_2} \rz_{\mu_4}\ri]\big|_{\text{connected}}\, \nonumber\\
&&+\mathbb{E}\le[\rz_{\mu_1} \rz_{\mu_4}\ri]\big|_{\text{connected}}\mathbb{E}\le[\rz_{\mu_2} \rz_{\mu_3}\ri]\big|_{\text{connected}}\, .\nonumber
\eea
Rearranging them in particular yields
\bea\label{C4}
&&\mathbb{E}\le[\rz_{\mu_1} \rz_{\mu_2}\rz_{\mu_3}\rz_{\mu_4}\ri]\big|_{\text{connected}}\, \\
&=&\mathbb{E}\le[\rz_{\mu_1} \rz_{\mu_2}\rz_{\mu_3}\rz_{\mu_4}\ri]\,\nonumber\\
&&-\mathbb{E}\le[\rz_{\mu_1} \rz_{\mu_2}\ri]\mathbb{E}\le[\rz_{\mu_3} \rz_{\mu_4}\ri]-\mathbb{E}\le[\rz_{\mu_1} \rz_{\mu_3}\ri]\mathbb{E}\le[\rz_{\mu_2} \rz_{\mu_4}\ri]-\mathbb{E}\le[\rz_{\mu_1} \rz_{\mu_4}\ri]\mathbb{E}\le[\rz_{\mu_2} \rz_{\mu_3}\ri]\, .\nonumber
\eea
If these examples do not suffice, here is yet another example to chew on:
\bea
\mathbb{E}\le[\rz_{\mu_1} \rz_{\mu_2}\rz_{\mu_3}\rz_{\mu_4}\rz_{\mu_5}\rz_{\mu_6}\ri]&=&\mathbb{E}\le[\rz_{\mu_1} \rz_{\mu_2}\rz_{\mu_3}\rz_{\mu_4}\rz_{\mu_5}\rz_{\mu_6}\ri]\big|_{\text{connected}}\, \\
&&+\mathbb{E}\le[\rz_{\mu_1} \rz_{\mu_2}\ri]\big|_{\text{connected}}\mathbb{E}\le[\rz_{\mu_3} \rz_{\mu_4}\ri]\big|_{\text{connected}}\mathbb{E}\le[\rz_{\mu_5} \rz_{\mu_6}\ri]\big|_{\text{connected}}\, \nonumber\\
&&+\le[14 \ \text{other}\ (2,2,2)\ \text{subdivisions}\ri]\, \nonumber\\
&&+\mathbb{E}\le[\rz_{\mu_1} \rz_{\mu_2}\rz_{\mu_3} \rz_{\mu_4}\ri]\big|_{\text{connected}}\mathbb{E}\le[\rz_{\mu_5} \rz_{\mu_6}\ri]\big|_{\text{connected}}\, \nonumber\\
&&+\le[14 \ \text{other}\ (4,2)\ \text{subdivisions}\ri]\, \nonumber
\eea
and hence
\bea
&&\mathbb{E}\le[\rz_{\mu_1} \rz_{\mu_2}\rz_{\mu_3}\rz_{\mu_4}\rz_{\mu_5}\rz_{\mu_6}\ri]\big|_{\text{connected}}\, \\
&=&\mathbb{E}\le[\rz_{\mu_1} \rz_{\mu_2}\rz_{\mu_3}\rz_{\mu_4}\rz_{\mu_5}\rz_{\mu_6}\ri]\,\nonumber\\
&&-\le\{\mathbb{E}\le[\rz_{\mu_1} \rz_{\mu_2}\rz_{\mu_3} \rz_{\mu_4}\ri]\mathbb{E}\le[\rz_{\mu_5}\rz_{\mu_6}\ri]+\le[14 \ \text{other}\ (4,2)\ \text{subdivisions}\ri]\ri\}\, \nonumber\\
&&+2\le\{\mathbb{E}\le[\rz_{\mu_1} \rz_{\mu_2}\ri]\mathbb{E}\le[\rz_{\mu_3} \rz_{\mu_4}\ri]\mathbb{E}\le[\rz_{\mu_5} \rz_{\mu_6}\ri]+\le[14 \ \text{other}\ (2,2,2)\ \text{subdivisions}\ri]\ri\}\, .\nonumber
\eea
We emphasize that these are just renderings of the definition~(\ref{cumu}). The power of this definition will be illustrated in the next two subsections. 

\subsection{Hierarchical clustering}\label{HC}
We often encounter situations with the hierarchy
\be\label{hier}
\mathbb{E}\le[\rz_{\mu_1} \cdots \rz_{\mu_m}\ri]\big|_{\text{connected}}=O(\epsilon^{\frac{m-2}{2}})\, 
\ee
where $\epsilon\ll 1$ is a small perturbative parameter and here again odd moments are assumed to vanish.
Often comes with the hierarchical structure is the asymptotic limit $\epsilon\rightarrow0$ where
\be\label{hier2}
\mathbb{E}\le[\rz_{\mu_1} \rz_{\mu_2}\ri]=\ker_{\mu_1 \mu_2}+\epsilon S_{\mu_1 \mu_2}+O(\epsilon^2)\, 
\ee
with the Gaussian kernel $\ker_{\mu_1 \mu_2}$ at zero $\epsilon$ and the leading self-energy correction $S_{\mu_1 \mu_2}$.
Let us also denote the leading four-point vertex
\be\label{hier3}
\mathbb{E}\le[\rz_{\mu_1} \rz_{\mu_2}\rz_{\mu_3}\rz_{\mu_4}\ri]\big|_{\text{connected}}=\epsilon \FPVA_{\mu_1 \mu_2 \mu_3 \mu_4}+O(\epsilon^2)\, . 
\ee
For instance this hierarchy holds for weakly-coupled field theories -- from which we are importing names such as self-energy, vertex, and metric -- and, in this paper, such hierarchical structure is inductively shown to hold for prior preactivations $\rvz^{(\ell)}$ with $\epsilon=\frac{1}{n_{\ell-1}}$ in the regime $n_1,\ldots,n_{L-1}\sim n\gg1$.
Note that, by definition, $\ker_{\mu_1 \mu_2}$ and $S_{\mu_1 \mu_2}$ are symmetric under $\mu_1\leftrightarrow\mu_2$ and $ \FPVA_{\mu_1 \mu_2 \mu_3 \mu_4}$ is symmetric under permutations of $(\mu_1,\mu_2,\mu_3,\mu_4)$.\footnote{In the main text the connected four-point preactivation correlation functions are symmetric under the permutations of four $(\text{sample},\text{neuron})$ indices, $\le\{(i_1,\alpha_1),(i_2, \alpha_2),(i_3,\alpha_3),(i_4,\alpha_4)\ri\}$.}

\subsection{Combinatorial hack}\label{hack}
So far we have reviewed the standard technology of Wick's contractions, connected correlation functions, and all that. Our objective now is to develop a method to evaluate $\mathbb{E}\le[\rz_{\mu_1} \cdots \rz_{\mu_m}\ri]$ for random variables obeying the hierarchical-clustering property~(\ref{hier}),\footnote{More precisely, we shall inductively use only the weaker proposition that $\mathbb{E}\le[\rz_{\mu_1} \cdots \rz_{\mu_m}\ri]\big|_{\text{connected}}=O(\epsilon^2)$ for $m\geq6$ along with Equations~(\ref{hier2}) and (\ref{hier3}).} which is the inductive hypothesis made in the main text; by extension, the resulting method (\ref{HclusteringDashDash}) lets us perturbatively evaluate $\mathbb{E}\le\{\mathcal{F}[\rvz]\ri\}$ for any function $\mathcal{F}$ that can be obtained as a limit of a sequence of analytic functions.

With the review of connected correlation functions passed us, first note that
\begin{align}
&\mathbb{E}\le[\rz_{\mu_1} \cdots \rz_{\mu_m}\ri]\, \label{Hclustering}\tag{CLUSTER}\\
=&\mathbb{E}\le[\rz_{\mu_1}\rz_{\mu_2}\ri]\cdots \mathbb{E}\le[\rz_{\mu_{m-1}}\rz_{\mu_m}\ri]+\le\{\le[(m-1)!!-1\ri]\ \text{other\ pairings}\ri\}\, \nonumber\\
&+\mathbb{E}\le[\rz_{\mu_1}\rz_{\mu_2}\rz_{\mu_3}\rz_{\mu_4}\ri]\big|_{\text{connected}} \mathbb{E}\le[\rz_{\mu_5}\rz_{\mu_6}\ri]\cdots\mathbb{E}\le[\rz_{\mu_{m-1}}\rz_{\mu_m}\ri] \,\nonumber\\
&+\le\{\le[{m \choose 4}\times (m-5)!!-1\ri]\ \text{other}\ (4,2,2,\ldots,2)\ \text{clusterings}\ri\}+O(\epsilon^2)\, \nonumber\\
=&\ker_{\mu_1 \mu_2}\cdots \ker_{\mu_{m-1} \mu_m}+\le\{\le[(m-1)!!-1\ri]\ \text{other\ pairings}\ri\}\, \nonumber\\
&+\epsilon S_{\mu_1 \mu_2}\ker_{\mu_3 \mu_4}\cdots\ker_{\mu_{m-1} \mu_m}\,\nonumber\\
&+\le\{\le[{m \choose 2}\times (m-3)!!-1\ri]\ \text{other\ such\ clusterings}\ri\}\, \nonumber\\
&+\epsilon \FPVA_{\mu_1 \mu_2 \mu_3 \mu_4}\ker_{\mu_5 \mu_6}\cdots\ker_{\mu_{m-1} \mu_m} \,\nonumber\\
&+\le\{\le[{m \choose 4}\times (m-5)!!-1\ri]\ \text{other}\ (4,2,2,\ldots,2)\ \text{clusterings}\ri\}+O(\epsilon^2)\, \nonumber\\
=&\le\langle \rz_{\mu_1} \cdots \rz_{\mu_m}\ri\rangle_{\ker}\,  \label{Hclusteringdash}\tag{CLUSTER'}\\
&+\epsilon S_{\mu_1 \mu_2}\le\langle \rz_{\mu_3} \cdots \rz_{\mu_m}\ri\rangle_{\ker}+\le\{\le[{m \choose 2}-1\ri]\ \text{other\ self-energy contractions}\ri\}\, \nonumber\\
&+\epsilon \FPVA_{\mu_1 \mu_2 \mu_3 \mu_4}\le\langle \rz_{\mu_5} \cdots \rz_{\mu_m}\ri\rangle_{\ker}+\le\{\le[{m \choose 4}-1\ri]\ \text{other\ vertex contractions}\ri\}+O(\epsilon^2)\, .\nonumber
\end{align}
where in the last equality Wick's theorem was used backward.

Below, let us use the inverse kernel $\le(\ker^{-1}\ri)^{\mu_1 \mu_2}$ as a metric to raise indices:
\bea
S^{\mu_1 \mu_2}&\equiv&\sum_{\mu'_1,\mu'_2}\le(\ker^{-1}\ri)^{\mu_1 \mu'_1}\le(\ker^{-1}\ri)^{\mu_2 \mu'_2}S_{\mu'_1 \mu'_2}\ \ \ \text{and}\, \\
\FPVA^{\mu_1 \mu_2 \mu_3 \mu_4}&\equiv&\sum_{\mu'_1,\ldots,\mu'_4}\le(\ker^{-1}\ri)^{\mu_1 \mu'_1}\cdots\le(\ker^{-1}\ri)^{\mu_4 \mu'_4}\FPVA_{\mu'_1 \mu'_2 \mu'_3 \mu'_4}\, .
\eea
Then, in order to simplify the second set of terms in Equation~(\ref{Hclusteringdash}) involving self-energy, note that
\bea
&& \le\langle \rz_{\mu_1} \cdots \rz_{\mu_m} \le(\sum_{\mu'_1,\mu'_2}S^{\mu'_1 \mu'_2} \rz_{\mu'_1}\rz_{\mu'_2}\ri)\ri\rangle_{\ker} \, \nonumber\\
&=&\sum_{\mu'_1,\mu'_2}S^{\mu'_1 \mu'_2} \Bigg[  \le\langle \rz_{\mu'_1}\rz_{\mu'_2}\ri\rangle_{\ker}\le\langle \rz_{\mu_1} \cdots \rz_{\mu_m}\ri\rangle_{\ker} \, \nonumber\\
&&\ \ \ \ \ \ \ \ \ \ \ \ \ \ \ + 2\le\langle \rz_{\mu_1}\rz_{\mu'_1}\ri\rangle_{\ker} \le\langle \rz_{\mu_2}\rz_{\mu'_2}\ri\rangle_{\ker}\le\langle \rz_{\mu_3} \cdots \rz_{\mu_m}\ri\rangle_{\ker}+\le\{\le[{m \choose 2}-1\ri]\ \text{other\ }\ (\mu_1,\mu_2)\ri\}\Bigg] \, \nonumber\\
&=&\le(\sum_{\mu'_1,\mu'_2}S^{\mu'_1 \mu'_2}\ker_{\mu'_1 \mu'_2}\ri)  \le\langle \rz_{\mu_1} \cdots \rz_{\mu_m}\ri\rangle_{\ker}\, \nonumber\\
&&+2S_{\mu_1 \mu_2} \le\langle \rz_{\mu_3} \cdots \rz_{\mu_m}\ri\rangle_{\ker}+\le\{\le[{m \choose 2}-1\ri]\ \text{other\ }\ (\mu_1,\mu_2)\ri\}\, \nonumber
\eea
where the symmetry $\mu_1\leftrightarrow\mu_2$ of $S_{\mu_1\mu_2}$ was used.
Hence, defining
\be\label{OSP}\tag{OS'}
\mathcal{O}_{S}[\rvz]\equiv \frac{1}{2}\sum_{\mu'_1,\mu'_2}S^{\mu'_1 \mu'_2}\le( \rz_{\mu'_1}\rz_{\mu'_2}-\ker_{\mu'_1 \mu'_2}\ri)\, ,
\ee
we obtain
\bea
&&\epsilon S_{\mu_1 \mu_2}\le\langle \rz_{\mu_3} \cdots \rz_{\mu_m}\ri\rangle_{\ker}+\le\{\le[{m \choose 2}-1\ri]\ \text{other\ }\ (\mu_1,\mu_2)\ri\}\, \\
&=&\epsilon\le\langle  \rz_{\mu_1} \cdots \rz_{\mu_m}\mathcal{O}_{S}[\rvz] \ri\rangle_{\ker}\, .
\eea
The similar algebraic exercise renders the other term in Equation~(\ref{Hclusteringdash}) to be 
\bea
&&\epsilon \FPVA_{\mu_1 \mu_2 \mu_3 \mu_4}\le\langle \rz_{\mu_5} \cdots \rz_{\mu_m}\ri\rangle_{\ker}+\le\{\le[{m \choose 4}-1\ri]\ \text{other\ }\ (\mu_1,\mu_2,\mu_3,\mu_4)\ri\}\, \\
&=&\epsilon\le\langle  \rz_{\mu_1} \cdots \rz_{\mu_m}\mathcal{O}_{V}[\rvz] \ri\rangle_{\ker}\, 
\eea
with
\be\label{OVP}\tag{OV'}
\mathcal{O}_{V}[\rvz]\equiv \frac{1}{24}\sum_{\mu'_1,\ldots,\mu'_4}\FPVA^{\mu'_1 \mu'_2 \mu'_3 \mu'_4} \le(\rz_{\mu'_1}\rz_{\mu'_2}\rz_{\mu'_3}\rz_{\mu'_4}-6\rz_{\mu'_1}\rz_{\mu'_2}\ker_{\mu'_3 \mu'_4}+3\ker_{\mu'_1 \mu'_2}\ker_{\mu'_3 \mu'_4}\ri)\, .
\ee

In summary, for any function $\mathcal{F}[\rvz]$ of random variables $\rz_{\mu}$
\be\label{HclusteringDashDash}\tag{HACK'}
\mathbb{E}\le\{\mathcal{F}[\rvz]\ri\}=\le\langle \mathcal{F}[\rvz]\ri\rangle_{\ker}+\epsilon\le\langle  \mathcal{F}[\rvz]\mathcal{O}_{S}[\rvz]\ri\rangle_{\ker}+\epsilon\le\langle  \mathcal{F}[\rvz]\mathcal{O}_{V}[\rvz]\ri\rangle_{\ker} +O(\epsilon^2)\, .
\ee

In order to get the expressions used in the main text at the $\ell$-th layer, we need only to replace $\mu\rightarrow(i,\alpha)$, identify $\epsilon=\frac{1}{n_{\ell-1}}$, and use the inductive hypotheses~(\ref{twopoint})
\be
\mathbb{E}\le[\rz_{i_1;\alpha_1} \rz_{i_2;\alpha_2}\ri]=\delta_{i_1 i_2} \le[\kert_{\alpha_1 \alpha_2}^{(\ell)}+\frac{1}{n_{\ell-1}}\SE^{(\ell)}_{\alpha_1 \alpha_2}+O\le(\frac{1}{n^{2}}\ri)\ri]\, \nonumber
\ee
and~(\ref{fourpoint})
\bea
&&\mathbb{E}\le[\rz_{i_1;\alpha_1} \rz_{i_2;\alpha_2}\rz_{i_3;\alpha_3}\rz_{i_4;\alpha_4}\ri]\big|_{\text{connected}}\, \nonumber\\
&=&\frac{1}{n_{\ell-1}} \bigg[\delta_{i_1 i_2}\delta_{i_3 i_4} \FPV^{(\ell)}_{(\alpha_1 \alpha_2) (\alpha_3 \alpha_4)}+\delta_{i_1 i_3}\delta_{i_2 i_4} \FPV^{(\ell)}_{(\alpha_1 \alpha_3) (\alpha_2 \alpha_4)}\, \nonumber\\
&&\ \ \ \ \ \ \ \ \ +\delta_{i_1 i_4}\delta_{i_2 i_3} \FPV^{(\ell)}_{(\alpha_1 \alpha_4) (\alpha_2 \alpha_3)}\bigg]+O\le(\frac{1}{n^{2}}\ri)\, . \nonumber
\eea
The operators in Equations~(\ref{OSP}) and~(\ref{OVP}) then become
\bea
\mathcal{O}_{S}[\rvz]&=& \frac{1}{2}\sum_{\alpha_1,\alpha_2} \SE_{(\ell)}^{\alpha_1 \alpha_2} \le[\le(\sum_{i=1}^{n_{\ell}} \rz_{i;\alpha_1}\rz_{i;\alpha_2}\ri)-n_{\ell}\kert^{(\ell)}_{\alpha_1 \alpha_2}\ri]\, \ \ \ \text{and}\,\nonumber\\
\mathcal{O}_{V}[\rvz]&=& \frac{1}{8}\sum_{\alpha_1,\alpha_2,\alpha_3,\alpha_4} \FPV_{(\ell)}^{(\alpha_1 \alpha_2) (\alpha_3 \alpha_4)}\, \nonumber\\
&&\times\Bigg[\le(\sum_{i=1}^{n_{\ell}} \rz_{i;\alpha_1}\rz_{i;\alpha_2}\ri)\le(\sum_{j=1}^{n_{\ell}} \rz_{j;\alpha_3}\rz_{j;\alpha_4}\ri)-2n_{\ell}\le(\sum_{i=1}^{n_{\ell}} \rz_{i;\alpha_1}\rz_{i;\alpha_2}\ri)\kert^{(\ell)}_{\alpha_3 \alpha_4}\, \nonumber\\
&&-4\le(\sum_{i=1}^{n_{\ell}} \rz_{i;\alpha_1}\rz_{i;\alpha_3}\ri)\kert^{(\ell)}_{\alpha_2 \alpha_4}+n_{\ell}^2\kert^{(\ell)}_{\alpha_1 \alpha_2}\kert^{(\ell)}_{\alpha_3 \alpha_4}+2n_{\ell}\kert^{(\ell)}_{\alpha_1 \alpha_3}\kert^{(\ell)}_{\alpha_2 \alpha_4}\Bigg]\, ,\nonumber
\eea
i.e., the operators in Equations~(\ref{OS}) and~(\ref{OV}) in the main text.

\newpage
\section{Full condensed proof}\label{activation_detail}
In this Appendix, we provide a full inductive proof for one of the main claims in the paper, streamlined in the main text. Namely, we assume at the $\ell$-th layer that Equations~(\ref{twopoint}) and~(\ref{fourpoint}) hold and that all the higher-point connected preactivation correlation functions are of order $O\le(\frac{1}{n^2}\ri)$ -- which are trivially true at $\ell=1$ -- and prove the same for the $(\ell+1)$-th layer. We assume the full mastery of Appendix~\ref{cumulants_for_dummies} or, conversely, this section can be used to test the mastery of Wick's tricks.

First, trivial Wick's contractions yield
\bea
&&\PCF^{(\ell+1)}_{i_1 \ldots i_{2m}; \alpha_1 \ldots \alpha_{2m}}\, \\
&=&\delta_{i_1 i_2}\cdots\delta_{i_{2m-1} i_{2m}}\sum_{k=0}^{m} \le[C_b^{(\ell+1)}\ri]^{m-k} \le[C_W^{(\ell+1)}\ri]^{k}\, \nonumber\\
&&\times\le\{\frac{1}{n_{\ell}^k}\sum_{j_1,\ldots,j_k=1}^{n_{\ell}}H^{(\ell)}_{j_1 j_1 \ldots j_k j_k; \alpha_1 \alpha_2 \ldots \alpha_{2k-1} \alpha_{2k}}+\le[{m \choose k}-1\ \text{others}\ri]\ri\}\, \nonumber\\
&&+\le[(2m-1)!!-1\ \text{other pairings}\ri]\, .\nonumber
\eea
Studiously disentangling cases with different numbers of repetitions in neuron indices $(j_1,\ldots,j_k)$, we notice that at order $O\le(\frac{1}{n}\ri)$, terms without repetition or with only one repetition contribute, finding
\bea
&&\frac{1}{n_{\ell}^k}\sum_{j_1,\ldots,j_k=1}^{n_{\ell}}H^{(\ell)}_{j_1 j_1 \ldots j_k j_k; \alpha_1 \alpha_2 \ldots \alpha_{2k-1} \alpha_{2k}}\, \\
&=&\le[\le\langle \sigma(\tilde{\rz}_{\alpha_1}) \sigma(\tilde{\rz}_{\alpha_2})\ri\rangle_{\kert^{(\ell)}}\cdots\le\langle \sigma(\tilde{\rz}_{\alpha_{2k-1}}) \sigma(\tilde{\rz}_{\alpha_{2k}})\ri\rangle_{\kert^{(\ell)}}\ri]\, \nonumber\\
&&+\frac{1}{n_{\ell}}\Bigg\{\le[\le\langle \sigma(\tilde{\rz}_{\alpha_1}) \sigma(\tilde{\rz}_{\alpha_2})\sigma(\tilde{\rz}_{\alpha_3})\sigma(\tilde{\rz}_{\alpha_4})\ri\rangle_{\kert^{(\ell)}}-\le\langle \sigma(\tilde{\rz}_{\alpha_1}) \sigma(\tilde{\rz}_{\alpha_2})\ri\rangle_{\kert^{(\ell)}}\le\langle \sigma(\tilde{\rz}_{\alpha_3}) \sigma(\tilde{\rz}_{\alpha_4})\ri\rangle_{\kert^{(\ell)}}\ri]\, \nonumber\\
&&\ \ \ \ \ \ \ \ \ \ \ \ \ \times\le[\le\langle \sigma(\tilde{\rz}_{\alpha_5}) \sigma(\tilde{\rz}_{\alpha_6})\ri\rangle_{\kert^{(\ell)}}\cdots\le\langle \sigma(\tilde{\rz}_{\alpha_{2k-1}}) \sigma(\tilde{\rz}_{\alpha_{2k}})\ri\rangle_{\kert^{(\ell)}}\ri]\, \nonumber\\
&&\ \ \ \ \ \ \ \ \ \ \ \ \ +\le[{k \choose 2}-1\ \text{others}\ri]\Bigg\}\, \nonumber\\
&&+\frac{1}{n_{\ell-1}}\le\langle \le[\sigma(\rz_{1;\alpha_1}) \sigma(\rz_{1;\alpha_2})\cdots\sigma(\rz_{k;\alpha_{2k-1}}) \sigma(\rz_{k;\alpha_{2k}})\ri]\le\{\mathcal{O}_{S}[\rvz]+\mathcal{O}_{V}[\rvz]\ri\}\ri\rangle_{\ker^{(\ell)}}\, \nonumber\\
&&+O\le(\frac{1}{n^2}\ri)\, \nonumber
\eea
where we used the inductive hierarchical assumption at the $\ell$-th layer, i.e., its consequence~(\ref{hackeq}) and denoted a single-neuron random vector $\tilde{\rvz}=\le\{\tilde{\rz}_{\alpha}\ri\}_{\alpha=1,\ldots,\ND}$ and the Gaussian integral with the core kernel $\le\langle \tilde{\rz}_{\alpha_1}\tilde{\rz}_{\alpha_2}\ri\rangle_{\kert^{(\ell)}}=\kert^{(\ell)}_{\alpha_1 \alpha_2}$. Plugging in expressions (\ref{OS},\ref{OV}) for operators $\mathcal{O}_{S,V}[\rvz]$,
\bea
&&\le\langle \le[\sigma(\rz_{1,\alpha_1}) \sigma(\rz_{1,\alpha_2})\cdots\sigma(\rz_{k,\alpha_{2k-1}}) \sigma(\rz_{k,\alpha_{2k}})\ri]\mathcal{O}_{S}[\rvz]\ri\rangle_{\ker^{(\ell)}}\, \\
&=&\frac{1}{2}\sum_{\alpha'_1,\alpha'_2}\SE_{(\ell)}^{\alpha'_1 \alpha'_2}\Bigg\{\le\langle \sigma(\tilde{\rz}_{\alpha_1}) \sigma(\tilde{\rz}_{\alpha_2})\le(\tilde{\rz}_{\alpha'_1}\tilde{\rz}_{\alpha'_2}-\kert^{(\ell)}_{\alpha'_1 \alpha'_2}\ri)\ri\rangle_{\kert^{(\ell)}}\, \nonumber\\
&&\ \ \ \ \ \ \ \ \ \ \ \ \ \ \ \ \ \ \ \ \ \ \ \ \times\le\langle \sigma(\tilde{\rz}_{\alpha_{3}}) \sigma(\tilde{\rz}_{\alpha_{4}})\ri\rangle_{\kert^{(\ell)}}\cdots\le\langle \sigma(\tilde{\rz}_{\alpha_{2k-1}}) \sigma(\tilde{\rz}_{\alpha_{2k}})\ri\rangle_{\kert^{(\ell)}}+\le[(k-1)\ \text{others}\ri]\Bigg\}\, \nonumber
\eea
and
\bea
&&\le\langle \le[\sigma(\rz_{1,\alpha_1}) \sigma(\rz_{1,\alpha_2})\cdots\sigma(\rz_{k,\alpha_{2k-1}}) \sigma(\rz_{k,\alpha_{2k}})\ri]\mathcal{O}_{V}[\rvz]\ri\rangle_{\ker^{(\ell)}}\, \\
&=&\frac{1}{8}\sum_{\alpha'_1,\alpha'_2,\alpha'_3,\alpha'_4}\FPV_{(\ell)}^{(\alpha'_1 \alpha'_2) (\alpha'_3 \alpha'_4)}\, \nonumber\\
&&\times\Bigg\{\Big\langle \sigma(\tilde{\rz}_{\alpha_1}) \sigma(\tilde{\rz}_{\alpha_2})\Big(\tilde{\rz}_{\alpha'_1}\tilde{\rz}_{\alpha'_2}\tilde{\rz}_{\alpha'_3}\tilde{\rz}_{\alpha'_4}-2\tilde{\rz}_{\alpha'_1}\tilde{\rz}_{\alpha'_2}\kert^{(\ell)}_{\alpha'_3 \alpha'_4}-4\tilde{\rz}_{\alpha'_1}\tilde{\rz}_{\alpha'_3}\kert^{(\ell)}_{\alpha'_2 \alpha'_4}\, \nonumber\\
&&\ \ \ \ \ \ \ \ \ \ \ \ \ \ \ \ \ \ \ \ \ \ \ \ \ \ \ \ \ \ \ \ \ \ +\kert^{(\ell)}_{\alpha'_1 \alpha'_2}\kert^{(\ell)}_{\alpha'_3 \alpha'_4}+2\kert^{(\ell)}_{\alpha'_1 \alpha'_3}\kert^{(\ell)}_{\alpha'_2 \alpha'_4}\Big)\Big\rangle_{\kert^{(\ell)}}\, \nonumber\\
&&\ \ \ \ \ \ \times\le\langle \sigma(\tilde{\rz}_{\alpha_{3}}) \sigma(\tilde{\rz}_{\alpha_{4}})\ri\rangle_{\kert^{(\ell)}}\cdots\le\langle \sigma(\tilde{\rz}_{\alpha_{2k-1}}) \sigma(\tilde{\rz}_{\alpha_{2k}})\ri\rangle_{\kert^{(\ell)}}+\le[(k-1)\ \text{others}\ri]\Bigg\}\, \nonumber\\
&&+\frac{1}{4}\sum_{\alpha'_1,\alpha'_2,\alpha'_3,\alpha'_4}\FPV_{(\ell)}^{(\alpha'_1 \alpha'_2) (\alpha'_3 \alpha'_4)}\, \nonumber\\
&&\times\Bigg\{\le\langle \sigma(\tilde{\rz}_{\alpha_1}) \sigma(\tilde{\rz}_{\alpha_2})\le(\tilde{\rz}_{\alpha'_1}\tilde{\rz}_{\alpha'_2}-\kert^{(\ell)}_{\alpha'_1 \alpha'_2}\ri)\ri\rangle_{\kert^{(\ell)}}\le\langle \sigma(\tilde{\rz}_{\alpha_3}) \sigma(\tilde{\rz}_{\alpha_4})\le(\tilde{\rz}_{\alpha'_3}\tilde{\rz}_{\alpha'_4}-\kert^{(\ell)}_{\alpha'_3 \alpha'_4}\ri)\ri\rangle_{\kert^{(\ell)}}\, \nonumber\\
&&\ \ \ \ \ \ \times\le\langle \sigma(\tilde{\rz}_{\alpha_{5}}) \sigma(\tilde{\rz}_{\alpha_{6}})\ri\rangle_{\kert^{(\ell)}}\cdots\le\langle \sigma(\tilde{\rz}_{\alpha_{2k-1}}) \sigma(\tilde{\rz}_{\alpha_{2k}})\ri\rangle_{\kert^{(\ell)}}+\le[{k \choose 2}-1\ \text{others}\ri]\Bigg\}\, \nonumber
\eea
As special cases, we obtain expressions advertised in the main text to be contained in this Appendix:
\bea
&&\widetilde{A}^{(\ell)}_{\alpha_1 \alpha_2}\equiv\frac{1}{n_{\ell}}\sum_{j=1}^{n_{\ell}}H_{j j;\alpha_1\alpha_2}^{(\ell)}\, \\
&=&\le\langle \sigma(\tilde{\rz}_{\alpha_1}) \sigma(\tilde{\rz}_{\alpha_2})\ri\rangle_{\kert^{(\ell)}}\, \nonumber\\
&&+\frac{1}{n_{\ell-1}}\Bigg[\frac{1}{2}\sum_{\alpha'_1,\alpha'_2} S_{(\ell)}^{\alpha'_1 \alpha'_2}\le\langle \sigma(\tilde{\rz}_{\alpha_1}) \sigma(\tilde{\rz}_{\alpha_2})(\tilde{\rz}_{\alpha'_1}\tilde{\rz}_{\alpha'_2}-\kert^{(\ell)}_{\alpha'_1 \alpha'_2})\ri\rangle_{\kert^{(\ell)}}\, \nonumber\\
&&\ \ \ \ \ \ \ \ \ \ \ \ +\frac{1}{8}\sum_{\alpha'_1,\alpha'_2,\alpha'_3,\alpha'_4} \FPV_{(\ell)}^{(\alpha'_1 \alpha'_2) (\alpha'_3 \alpha'_4)}\Big\langle \sigma(\tilde{\rz}_{\alpha_1}) \sigma(\tilde{\rz}_{\alpha_2})\,\nonumber\\
&&\ \ \ \ \ \ \ \ \ \ \ \ \times\Big(\tilde{\rz}_{\alpha'_1}\tilde{\rz}_{\alpha'_2}\tilde{\rz}_{\alpha'_3}\tilde{\rz}_{\alpha'_4}-2\tilde{\rz}_{\alpha'_1}\tilde{\rz}_{\alpha'_2}\kert^{(\ell)}_{\alpha'_3 \alpha'_4}-4\tilde{\rz}_{\alpha'_1}\tilde{\rz}_{\alpha'_3}\kert^{(\ell)}_{\alpha'_2 \alpha'_4}\, \nonumber\\
&&\ \ \ \ \ \ \ \ \ \ \ \ +\kert^{(\ell)}_{\alpha'_1 \alpha'_2}\kert^{(\ell)}_{\alpha'_3 \alpha'_4}+2\kert^{(\ell)}_{\alpha'_1 \alpha'_3}\kert^{(\ell)}_{\alpha'_2 \alpha'_4}\Big)\Big\rangle_{\kert^{(\ell)}}\Bigg]+O\le(\frac{1}{n^2}\ri)\, \nonumber
\eea
and
\bea
&&\widetilde{B}^{(\ell)}_{(\alpha_1 \alpha_2) (\alpha_3 \alpha_4)}\equiv\frac{1}{n^2_{\ell}}\sum_{j_1,j_2=1}^{n_{\ell}}\le[\ACF^{(\ell)}_{j_1 j_1 j_2 j_2; \alpha_1 \alpha_2 \alpha_3 \alpha_4}-\ACF^{(\ell)}_{j_1 j_1; \alpha_1 \alpha_2}\ACF^{(\ell)}_{j_2 j_2; \alpha_3 \alpha_4}\ri]\, \\
&=&\frac{1}{n_{\ell}}\Bigg\{\le\langle \sigma(\tilde{\rz}_{\alpha_1}) \sigma(\tilde{\rz}_{\alpha_2}) \sigma(\tilde{\rz}_{\alpha_3}) \sigma(\tilde{\rz}_{\alpha_4})\ri\rangle_{\kert^{(\ell)}}-\le\langle \sigma(\tilde{\rz}_{\alpha_1}) \sigma(\tilde{\rz}_{\alpha_2})\ri\rangle_{\kert^{(\ell)}}\le\langle \sigma(\tilde{\rz}_{\alpha_3}) \sigma(\tilde{\rz}_{\alpha_4})\ri\rangle_{\kert^{(\ell)}}\,\nonumber\\
&&\ \ \ \ \ \ \ +\frac{1}{4}\le(\frac{n_{\ell}}{n_{\ell-1}}\ri)\sum_{\alpha'_1,\alpha'_2,\alpha'_3,\alpha'_4} \FPV_{(\ell)}^{(\alpha'_1 \alpha'_2) (\alpha'_3 \alpha'_4)}\le\langle \sigma(\tilde{\rz}_{\alpha_1}) \sigma(\tilde{\rz}_{\alpha_2})(\tilde{\rz}_{\alpha'_1}\tilde{\rz}_{\alpha'_2}-\kert^{(\ell)}_{\alpha'_1 \alpha'_2})\ri\rangle_{\kert^{(\ell)}}\,\nonumber\\
&&\ \ \ \ \ \ \ \ \ \ \ \ \ \ \ \ \ \ \ \ \ \ \ \times\le\langle \sigma(\tilde{\rz}_{\alpha_3}) \sigma(\tilde{\rz}_{\alpha_4})(\tilde{\rz}_{\alpha'_3}\tilde{\rz}_{\alpha'_4}-\kert^{(\ell)}_{\alpha'_3 \alpha'_4})\ri\rangle_{\kert^{(\ell)}}\Bigg\}+O\le(\frac{1}{n^2}\ri)\, .\nonumber
\eea

Assembling everything,
\bea
&&\PCF^{(\ell+1)}_{i_1 \ldots i_{2m};\alpha_1 \ldots \alpha_{2m}}\, \\
&=&\delta_{i_1 i_2}\cdots\delta_{i_{2m-1} i_{2m}}\prod_{k=1}^{m}\le[C_b^{(\ell+1)}+C_W^{(\ell+1)}\widetilde{A}^{(\ell)}_{\alpha_{2k-1} \alpha_{2k}}\ri]\, \nonumber\\
&&+\le[(2m-1)!!-1\ \text{other pairings}\ri]\, \nonumber\\
&&+\delta_{i_1 i_2}\cdots\delta_{i_{2m-1} i_{2m}}\widetilde{B}^{(\ell)}_{(\alpha_1 \alpha_2) (\alpha_3 \alpha_4)}\prod_{k=3}^{m}\le[C_b^{(\ell+1)}+C_W^{(\ell+1)}\widetilde{A}^{(\ell)}_{\alpha_{2k-1} \alpha_{2k}}\ri]\, \nonumber\\
&&+\le\{\le[3\times{2m \choose 4}\times (2m-5)!!-1\ri]\ \text{other}\ (4,2,2,\ldots,2)\ \text{clusterings}\ri\}\, \nonumber\\
&&+O\le(\frac{1}{n^2}\ri)\, .\nonumber
\eea
In particular,
\bea
\PCF^{(\ell+1)}_{i_1 i_2; \alpha_1 \alpha_2}&=&\delta_{i_1 i_2} \le[C_b^{(\ell+1)}+C_W^{(\ell+1)}\widetilde{A}^{(\ell)}_{\alpha_{1} \alpha_{2}}\ri]+O\le(\frac{1}{n^2}\ri)\, ,\\
\PCF^{(\ell+1)}_{i_1 i_2 i_3 i_4; \alpha_1 \alpha_2 \alpha_3 \alpha_4}\Big\vert_{\text{connected}}\, &=&\delta_{i_1 i_2}\delta_{i_3 i_4}\widetilde{B}^{(\ell)}_{(\alpha_1 \alpha_2) (\alpha_3 \alpha_4)}+\delta_{i_1 i_3}\delta_{i_2 i_4}\widetilde{B}^{(\ell)}_{(\alpha_1 \alpha_3) (\alpha_2 \alpha_4)}\, \nonumber\\
&&+\delta_{i_1 i_4}\delta_{i_2 i_3}\widetilde{B}^{(\ell)}_{(\alpha_1 \alpha_4) (\alpha_2 \alpha_3)}+O\le(\frac{1}{n^2}\ri)\, ,\ \ \ \text{and}\, \nonumber\\
\PCF^{(\ell+1)}_{i_1 i_2 \ldots i_{2m-1} i_{2m};\alpha_1 \alpha_2 \ldots \alpha_{2m-1} \alpha_{2m}}\Big\vert_{\text{connected}}&=&O\le(\frac{1}{n^2}\ri)\, ,\ \ \ \text{for}\ \ \ 2m\geq6\, .
\eea
completing our inductive proof. Note that $\widetilde{B}^{(\ell)}_{(\alpha_1 \alpha_2) (\alpha_3 \alpha_4)}=O\le(\frac{1}{n}\ri)$.

Nowhere in our derivation had we assumed anything about the form of activation functions. The only potential exceptions to our formalism are exponentially growing activation functions -- which we never see in practice -- that would make the Gaussian integrals unintegrable.

\newpage
\section{Bestiary of concrete examples}\label{bestiary}
\subsection{Quadratic activation}\label{bestiary_quad}
Let us take multilayer perceptrons with quadratic activation, $\sigma(z)=z^2$, and study the distributions of preactivations in the second layer as another illustration of our technology. From the master recursion relations~(\ref{R1}-\ref{R3}) with the initial condition~(\ref{initial}), Wick's contractions yield
\begin{align}
\kert^{(2)}_{\alpha_1 \alpha_2}=&C_b^{(2)}+C_W^{(2)}\le[\kert^{(1)}_{\alpha_1 \alpha_1}\kert^{(1)}_{\alpha_2 \alpha_2}+2\kert^{(1)}_{\alpha_1 \alpha_2}\kert^{(1)}_{\alpha_1 \alpha_2}\ri]\, ,\label{R1quad}\\
\frac{\FPV^{(2)}_{(\alpha_1 \alpha_2) (\alpha_3 \alpha_4)}}{\le[C_W^{(2)}\ri]^2}=&2\bigg[\kert^{(1)}_{\alpha_1 \alpha_1}\kert^{(1)}_{\alpha_3 \alpha_3}\le(\kert^{(1)}_{\alpha_2 \alpha_4}\ri)^2+\kert^{(1)}_{\alpha_1 \alpha_1}\kert^{(1)}_{\alpha_4 \alpha_4}\le(\kert^{(1)}_{\alpha_2 \alpha_3}\ri)^2\, \label{R2quad}\\
&\ \ \ +\kert^{(1)}_{\alpha_2 \alpha_2}\kert^{(1)}_{\alpha_3 \alpha_3}\le(\kert^{(1)}_{\alpha_1 \alpha_4}\ri)^2+\kert^{(1)}_{\alpha_2 \alpha_2}\kert^{(1)}_{\alpha_4 \alpha_4}\le(\kert^{(1)}_{\alpha_1 \alpha_3}\ri)^2\bigg]\, \nonumber\\
&+4\le[\le(\kert^{(1)}_{\alpha_1 \alpha_3}\ri)^2\le(\kert^{(1)}_{\alpha_2 \alpha_4}\ri)^2+\le(\kert^{(1)}_{\alpha_1 \alpha_4}\ri)^2\le(\kert^{(1)}_{\alpha_2 \alpha_3}\ri)^2\ri]\, \nonumber\\
&+8\bigg[\kert^{(1)}_{\alpha_1 \alpha_1}\kert^{(1)}_{\alpha_2 \alpha_3}\kert^{(1)}_{\alpha_3 \alpha_4}\kert^{(1)}_{\alpha_4 \alpha_2}+\kert^{(1)}_{\alpha_2 \alpha_2}\kert^{(1)}_{\alpha_3 \alpha_4}\kert^{(1)}_{\alpha_4 \alpha_1}\kert^{(1)}_{\alpha_1 \alpha_3}\, \nonumber\\
&\ \ \ \ \ \ +\kert^{(1)}_{\alpha_3 \alpha_3}\kert^{(1)}_{\alpha_4 \alpha_1}\kert^{(1)}_{\alpha_1 \alpha_2}\kert^{(1)}_{\alpha_2 \alpha_4}+\kert^{(1)}_{\alpha_4 \alpha_4}\kert^{(1)}_{\alpha_1 \alpha_2}\kert^{(1)}_{\alpha_2 \alpha_3}\kert^{(1)}_{\alpha_3 \alpha_1}\bigg]\, \nonumber\\
&+16\le[\kert^{(1)}_{\alpha_1 \alpha_2}\kert^{(1)}_{\alpha_1 \alpha_3}\kert^{(1)}_{\alpha_2 \alpha_4}\kert^{(1)}_{\alpha_3 \alpha_4}+\kert^{(1)}_{\alpha_1 \alpha_2}\kert^{(1)}_{\alpha_1 \alpha_4}\kert^{(1)}_{\alpha_2 \alpha_3}\kert^{(1)}_{\alpha_3 \alpha_4}\ri]\, \nonumber\\
&+16\kert^{(1)}_{\alpha_1 \alpha_3}\kert^{(1)}_{\alpha_1 \alpha_4}\kert^{(1)}_{\alpha_2 \alpha_3}\kert^{(1)}_{\alpha_2 \alpha_4}\, ,\ \ \ \text{and}\nonumber\\
\SE^{(2)}_{\alpha_1 \alpha_2}=&0\, .\label{R3quad}
\end{align}
where $\kert^{(1)}_{\alpha_1 \alpha_2}=C_{b}^{(1)}+C_{W}^{(1)}\cdot\le(\frac{\inp_{\alpha_1}\cdot\inp_{\alpha_2}}{n_0}\ri)$. These expressions are used in the main text
for the experimental study of finite-width corrections on Bayesian inference.

\subsection{Details for single-input cases}\label{bestiary_mono}
The recursive relations simplify drastically for the case of a single input, $\ND=1$. Setting $C_{b}^{(\ell)}=0$ for simplicity and dropping $\alpha$ index, our recursive equations reduce to
\begin{align}\label{R1single}
\kert^{(\ell+1)}=&C^{(\ell+1)}_{W} \le\langle \le[\sigma(\tilde{\rz})\ri]^2\ri\rangle_{\kert^{(\ell)}}\, ,\\
\frac{\FPV^{(\ell+1)}}{\le(\kert^{(\ell+1)}\ri)^2}=&\le(\frac{\le\langle \le[\sigma(\tilde{\rz})\ri]^4\ri\rangle_{\kert^{(\ell)}}}{\le\langle \le[\sigma(\tilde{\rz})\ri]^2\ri\rangle_{\kert^{(\ell)}}^2}-1\ri)+\frac{1}{4}\le(\frac{n_{\ell}}{n_{\ell-1}}\ri)\le(\frac{\le\langle \le[\sigma(\tilde{\rz})\ri]^2\tilde{\rz}^2\ri\rangle_{\kert^{(\ell)}}}{\le\langle \le[\sigma(\tilde{\rz})\ri]^2\ri\rangle_{\kert^{(\ell)}}\kert^{(\ell)}}-1\ri)^2\cdot\frac{\FPV^{(\ell)}}{\le(\kert^{(\ell)}\ri)^2}\, , \ \text{and}\,\label{R2single}\\
\frac{\SE^{(\ell+1)}}{\kert^{(\ell+1)}}=&\frac{1}{2}\le(\frac{n_{\ell}}{n_{\ell-1}}\ri)\le(\frac{\le\langle \le[\sigma(\tilde{\rz})\ri]^2\tilde{\rz}^2\ri\rangle_{\kert^{(\ell)}}}{\le\langle \le[\sigma(\tilde{\rz})\ri]^2\ri\rangle_{\kert^{(\ell)}}\kert^{(\ell)}}-1\ri)\cdot\frac{\SE^{(\ell)}}{\kert^{(\ell)}}\, \label{R3single}\\
&+\frac{1}{8}\le(\frac{n_{\ell}}{n_{\ell-1}}\ri)\le(\frac{\le\langle \le[\sigma(\tilde{\rz})\ri]^2\tilde{\rz}^4\ri\rangle_{\kert^{(\ell)}}}{\le\langle \le[\sigma(\tilde{\rz})\ri]^2\ri\rangle_{\kert^{(\ell)}}\le(\kert^{(\ell)}\ri)^2}-6\frac{\le\langle \le[\sigma(\tilde{\rz})\ri]^2\tilde{\rz}^2\ri\rangle_{\kert^{(\ell)}}}{\le\langle \le[\sigma(\tilde{\rz})\ri]^2\ri\rangle_{\kert^{(\ell)}}\kert^{(\ell)}}+3\ri)\cdot\frac{\FPV^{(\ell)}}{\le(\kert^{(\ell)}\ri)^2}\, .\nonumber
\end{align}

\subsubsection{Monomials with single input}\label{monomial_single}
For monomial activations, $\sigma(z)=z^p$, such as in deep linear networks~\cite{SMG2013} and quadratic activations~\cite{LMZ2017},
\begin{align}
\kert^{(\ell+1)}&=\le[(2p-1)!! C_W^{(\ell+1)} \ri]\le(\kert^{(\ell)}\ri)^p\, ,\\
\frac{\FPV^{(\ell+1)}}{\le(\kert^{(\ell+1)}\ri)^2}&=\le\{\frac{(4p-1)!!}{\le[(2p-1)!!\ri]^2}-1\ri\}+p^2\le(\frac{n_{\ell}}{n_{\ell-1}}\ri)\frac{\FPV^{(\ell)}}{\le(\kert^{(\ell)}\ri)^2}\, ,\ \ \ \text{and}\, \\
\frac{\SE^{(\ell+1)}}{\kert^{(\ell+1)}}&=\le(\frac{n_{\ell}}{n_{\ell-1}}\ri)\le[p\frac{\SE^{(\ell)}}{\kert^{(\ell)}}+\frac{p(p-1)}{2}\frac{\FPV^{(\ell)}}{\le(\kert^{(\ell)}\ri)^2}\ri]\, .
\end{align}
In particular the four-point vertex solution is given by
\be
\frac{1}{n_{\ell-1}p^{2(\ell-1)}}\frac{\FPV^{(\ell)}}{\le(\kert^{(\ell)}\ri)^2}=\le\{\frac{(4p-1)!!}{\le[(2p-1)!!\ri]^2}-1\ri\}\le(\sum_{\ell'=1}^{\ell-1}\frac{1}{n_{\ell'}p^{2\ell'}}\ri)\, .
\ee
The factor $\le(\sum_{\ell'}\frac{1}{n_{\ell'}p^{2\ell'}}\ri)$ generalizes the factor $\le(\sum_{\ell'}\frac{1}{n_{\ell'}}\ri)$ for linear and ReLU activations. Following Ref.~\cite{HR2018}, this factor guides us to narrow hidden layers as we pass through nonlinear activations for $p>1$.

\subsubsection{ReLU with single input}\label{reluagain_single}
ReLU activation, $\sigma(z)=\mathrm{max}(0,z)$, can also be worked out for a single input through Wick's contractions, noting that the Gaussian integral is halved, yielding
\begin{align}
\kert^{(\ell+1)}&=\le[\frac{C^{(\ell+1)}_W}{2} \ri]\kert^{(\ell)}\, , \\
\frac{\FPV^{(\ell+1)}}{\le(\kert^{(\ell+1)}\ri)^2}&=5+\le(\frac{n_{\ell}}{n_{\ell-1}}\ri)\frac{\FPV^{(\ell)}}{\le(\kert^{(\ell)}\ri)^2}\, ,\ \ \ \text{and}\, \\
\frac{\SE^{(\ell+1)}}{\kert^{(\ell+1)}}&=\le(\frac{n_{\ell}}{n_{\ell-1}}\ri)\frac{\SE^{(\ell)}}{\kert^{(\ell)}}\, .
\end{align}
Setting $C^{(\ell)}_W=2$ for simplicity, these equations can be solved, leading to
\begin{align}
\kert^{(\ell)}&=\kert^{(1)}=\frac{\vert\vert\inp\vert|_2^2}{n_0}\, , \\
\frac{1}{n_{\ell-1}}\FPV^{(\ell)}&=5\le(\sum_{\ell'=1}^{\ell-1}\frac{1}{n_{\ell'}}\ri)\le(\kert^{(1)}\ri)^2\, ,\ \ \ \text{and}\, \\
\SE^{(\ell)}&=0\, .
\end{align}

\subsection{More experiments on output distributions}\label{more_experiments}
\begin{figure}[h]
\centering{
 \includegraphics[width=0.35\linewidth]{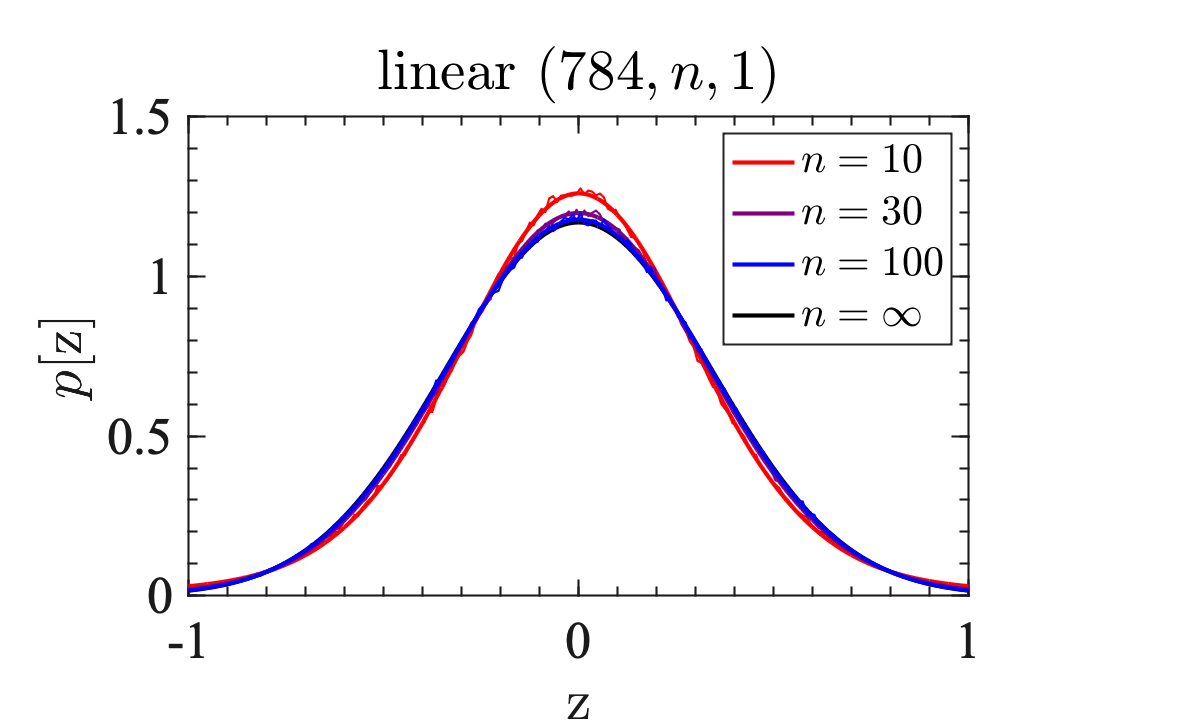}\hspace{-0.8cm}
\includegraphics[width=0.35\linewidth]{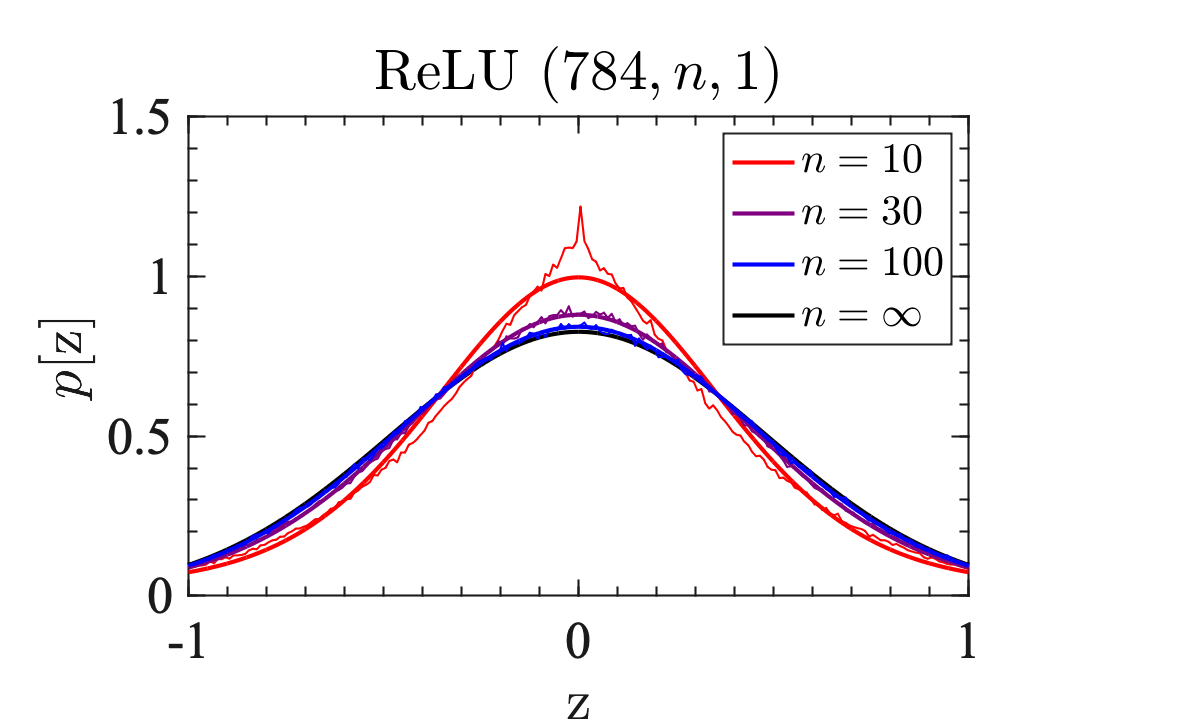}\hspace{-0.8cm}
 \includegraphics[width=0.35\linewidth]{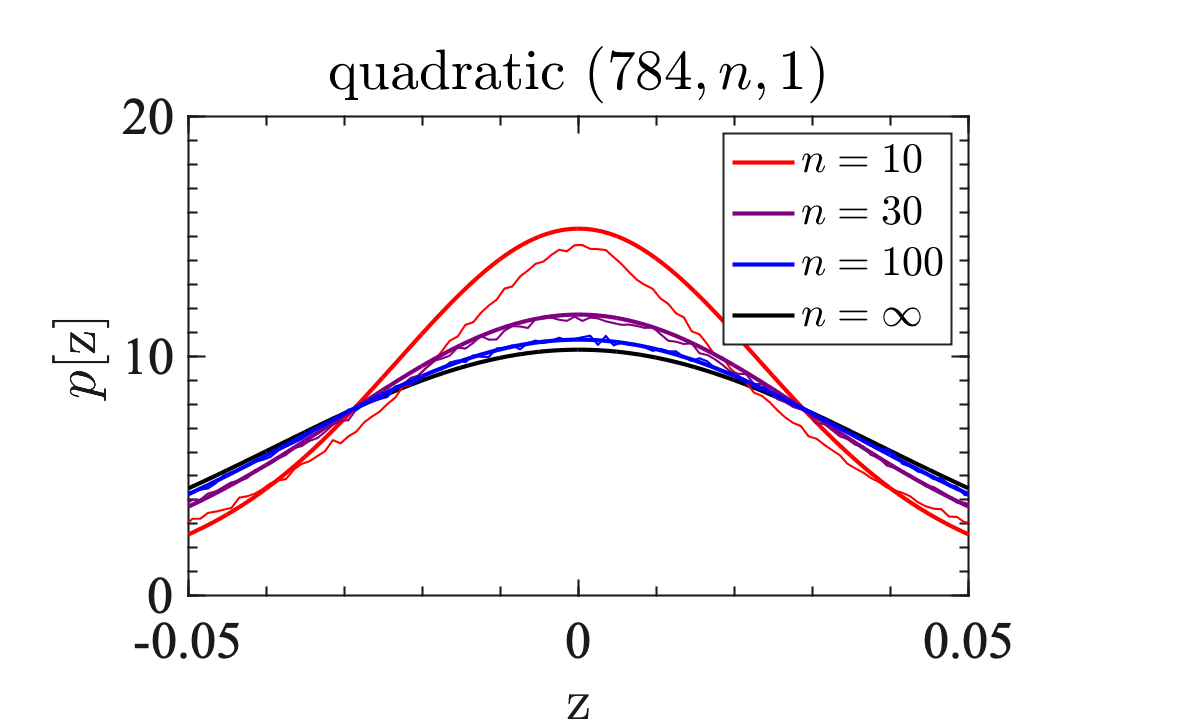}
}
\centering{
 \includegraphics[width=0.35\linewidth]{doublelinear.png}\hspace{-0.8cm}
\includegraphics[width=0.35\linewidth]{doubleReLU.png}\hspace{-0.8cm}
 \includegraphics[width=0.35\linewidth]{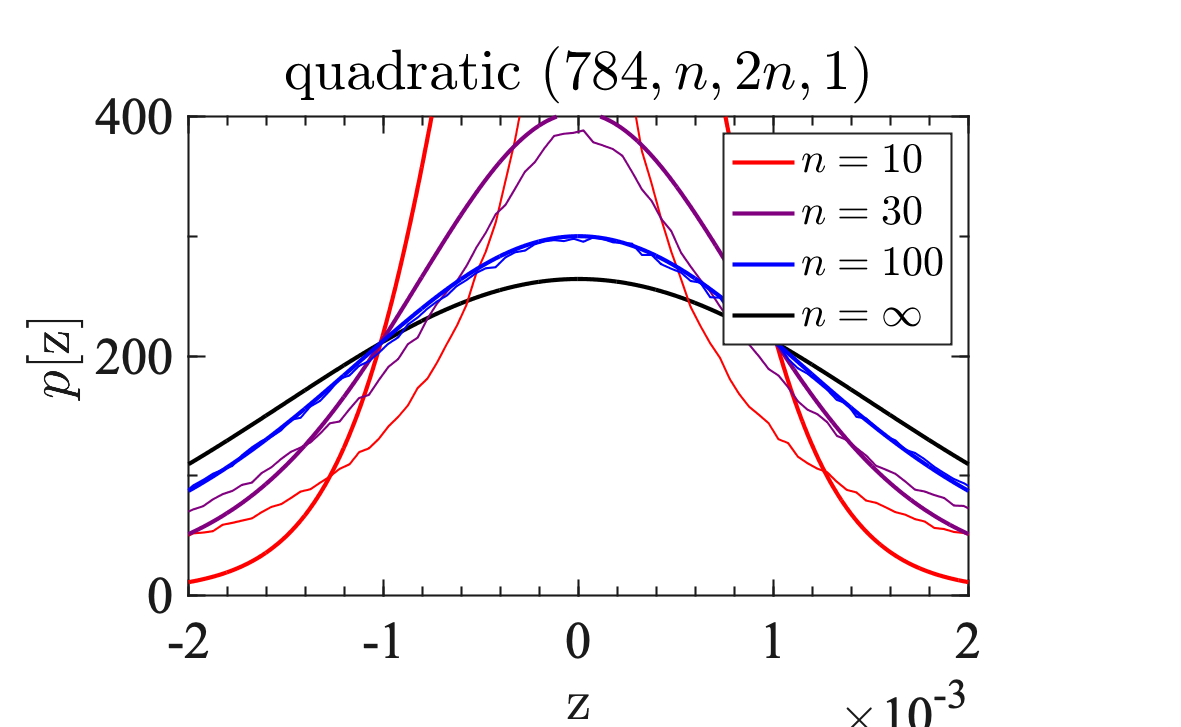}
}
\caption{Comparison between theory and experiments for prior distributions of outputs for a single input. Our theoretical predictions (smooth thick lines) and experimental data (rugged thin lines) agree, correctly capturing the initial deviations from the Gaussian processes (black, $n=\infty$), at least down to $n=n_{\star}$ with $n_{\star}\sim10$ for linear cases, $n_{\star}\sim30$ for ReLU cases and depth $L=2$ quadratic case, and $n_{\star}\sim100$ for depth $L=3$ quadratic case. This also illustrates that nonlinear activations quickly amplify non-Gaussianity.}
\label{NNNGP1ex}
\end{figure}

Here is an extended version of experiments in Section~\ref{Fexperiment}. As in the main text, take a single black-white image of hand-written digits from the MNIST dataset as an $n_0=784$-dimensional input, without preprocessing. Set bias variance $C^{(\ell)}_{b}=0$, weight variance $C^{(\ell)}_W=C_W$,  and use activations $\sigma(z)=z$ (linear) with $C_W=1$, $\sigma(z)=z^2$ (quadratic) with $C_W=\frac{1}{3}$, and $\sigma(z)=\mathrm{max}(0,z)$ (ReLU) with $C_W=2$. For all three cases, we consider both depth $L=2$ with widths $(n_0,n_1,n_2)=(784,n,1)$ and depth $L=3$ with widths $(n_0,n_1,n_2,n_3)=(784,n,2n,1)$. As in Figure~\ref{NNNGP1}, in Figure~\ref{NNNGP1ex}, for each width-parameter $n$ of the hidden layers we record the prior distribution of outputs over $10^6$ instances of Gaussian weights and compare it with the theoretical prediction. Results again corroborate our theory.

\newpage
\

\newpage
\section{Finite-width corrections on Bayesian inference}\label{NNNGPmanipulation}
In order to massage Equation~(\ref{NNNGPP}) into an actionable form, first playing with the metric inversions and defining $\overline{\phi}_{i}^{\ \alpha}\equiv\sum_{\alpha'}(\kert^{-1})^{\alpha \alpha'}\overline{\phi}_{i;\alpha'}$, the mean prediction becomes
\begin{align}
&\le(\GPM\ri)_{i;\tes}+\epsilon \sum_{\alpha_1,\tes_1,\alpha_0}\le(\kert_{\Delta}\ri)_{\tes \tes_1}\overline{\phi}_{i}^{\ \alpha_1}\le(\kert^{-1}\ri)^{\tes_1 \alpha_0}\,  \label{NNNGPP1}\\
&\ \ \ \ \ \ \ \ \ \ \ \ \times\Bigg\{\SE^{(L)}_{\alpha_0 \alpha_1}-\sum_{\alpha_2,\alpha_3}\le[\FPV^{(L)}_{(\alpha_0 \alpha_2) (\alpha_1 \alpha_3)}+\frac{n_L}{2}\FPV^{(L)}_{(\alpha_0 \alpha_1) (\alpha_2 \alpha_3)}\ri]\, \nonumber\\
&\ \ \ \ \ \ \ \ \ \ \ \  \ \ \ \ \ \ \times\le[\le(\kert^{-1}\ri)^{\alpha_2 \alpha_3}-\le(\kert^{-1}\ri)^{\alpha_2 \tes_2}\le(\kert_{\Delta}\ri)_{\tes_2 \tes_3}\le(\kert^{-1}\ri)^{\tes_3 \alpha_3}\ri]\, \nonumber\\
&\ \ \ \ \ \ \ \ \ \ \ \ \ \ \ \ \ \ \ +\frac{1}{2}\sum_{\alpha_2,\alpha_3}\FPV^{(L)}_{(\alpha_0 \alpha_1) (\alpha_2 \alpha_3)}\le(\sum_{j}\overline{\phi}_{j}^{\ \alpha_2}\overline{\phi}_{j}^{\ \alpha_3}\ri)\Bigg\}\, .\nonumber
\end{align}
This expression simplifies drastically through the identity
\be
\begin{pmatrix}
\kert_{\text{RR}} & \kert_{\text{RE}}\\
\kert_{\text{ER}}  & \kert_{\text{EE}}
\end{pmatrix}^{-1}
=
\begin{pmatrix}
\kert_{\text{RR}}^{-1}+\kert_{\text{RR}}^{-1}\kert_{\text{RE}}\kert_{\Delta}^{-1}\kert_{\text{ER}}\kert_{\text{RR}}^{-1} & -\kert_{\text{RR}}^{-1}\kert_{\text{RE}}\kert_{\Delta}^{-1}\\
-\kert_{\Delta}^{-1}\kert_{\text{ER}}\kert_{\text{RR}}^{-1}  & \kert_{\Delta}^{-1}
\end{pmatrix}
\, ,
\ee
which can be checked explicitly, recalling $\kert_{\Delta}\equiv \kert_{\text{EE}}- \kert_{\text{ER}} \kert_{\text{RR}}^{-1} \kert_{\text{RE}}$. Incidentally, this identity can also be used to prove Equation~(\ref{GPH}). Now equipped with this identity, recalling $\overline{\phi}_{i;\alpha}\equiv [\le(y_{\text{R}}\ri)_{i;\tra},\le(\GPM\ri)_{i;\tes}]$, we notice that $\overline{\phi}_{i}^{\ \tra}=\sum_{\tra'}\le(\kert_{\text{RR}}^{-1}\ri)^{\tra \tra'}\le(y_{\text{R}}\ri)_{i;\tra'}$ and $\overline{\phi}_{i}^{\ \tes_1}=0$. Similarly
\be
\le[\le(\kert^{-1}\ri)^{\tra_2 \tra_3}-\sum_{\tes_2,\tes_3}\le(\kert^{-1}\ri)^{\tra_2 \tes_2}\le(\kert_{\Delta}\ri)_{\tes_2 \tes_3}\le(\kert^{-1}\ri)^{\tes_3 \tra_3}\ri]=\le(\kert_{\text{RR}}^{-1}\ri)^{\tra_2 \tra_3}\, \nonumber
\ee
and other components [i.e.~with one or both of training components $(\tra_2,\tra_3)$ replaced by test components $\tes$] vanish.
Equation~(\ref{NNNGPP1}) thus simplifies to
\begin{align}
&\le(\GPM\ri)_{i;\tes}+\epsilon \sum_{\tra_1,\tes_1,\alpha_0}\le(\kert_{\Delta}\ri)_{\tes \tes_1}\overline{\phi}_{i}^{\ \tra_1}\le(\kert^{-1}\ri)^{\tes_1 \alpha_0}\Bigg[\SE^{(L)}_{\alpha_0 \tra_1}+\frac{1}{2}\sum_{\tra_2,\tra_3}\FPV^{(L)}_{(\alpha_0 \tra_1) (\tra_2 \tra_3)}\le(\sum_{j}\overline{\phi}_{j}^{\ \tra_2}\overline{\phi}_{j}^{\ \tra_3}\ri)\, \label{NNNGPP2}\\
&\ \ \ \ \ \ \ \ \ \ \ \  \ \ \ \ \ \ -\sum_{\tra_2,\tra_3}\le(\FPV^{(L)}_{(\alpha_0 \tra_2) (\tra_1 \tra_3)}+\frac{n_L}{2}\FPV^{(L)}_{(\alpha_0 \tra_1) (\tra_2 \tra_3)}\ri)\le(\kert_{\text{RR}}^{-1}\ri)^{\tra_2 \tra_3}\, \Bigg]\, \nonumber
\end{align}
Finally, denoting the matrix inside the parenthesis to be
\begin{align}\label{NNNGPPdash}\tag{NGPM'}
A_{\alpha_0 \tra_1}\equiv& S^{(L)}_{\alpha_0 \tra_1}+\frac{1}{2}\sum_{\tra_2,\tra_3,\tra'_2,\tra'_3}\FPV^{(L)}_{(\alpha_0 \tra_1) (\tra_2 \tra_3)}\le(\sum_{j}\overline{\phi}_{j}^{\ \tra_2}\overline{\phi}_{j}^{\ \tra_3}\ri)\, \\
&-\sum_{\tra_2,\tra_3}\le(\FPV^{(L)}_{(\alpha_0 \tra_2) (\tra_1 \tra_3)}+\frac{n_L}{2}\FPV^{(L)}_{(\alpha_0 \tra_1) (\tra_2 \tra_3)}\ri)\le(\kert_{\text{RR}}^{-1}\ri)^{\tra_2 \tra_3}\, ,\nonumber
\end{align}
and noticing $ \sum_{\tes_1}\le(\kert_{\Delta}\ri)_{\tes \tes_1}\le(\kert^{-1}\ri)^{\tes_1 \tra_0}=-\le(\kert_{\text{ER}}\kert^{-1}_{\text{RR}}\ri)_{\tes}^{\ \ \ \tra_0}$ and $ \sum_{\tes_1}\le(\kert_{\Delta}\ri)_{\tes \tes_1}\le(\kert^{-1}\ri)^{\tes_1 \tes_0}=\delta_{\tes}^{\ \tes_0}$,
\begin{align}\label{NNNGPPdashdash}\tag{NGPM''}
&\le(\GPM\ri)_{i; \tes}+\epsilon \sum_{\tra_1}\overline{\phi}_{i}^{\ \tra_1}\le[A_{\tes \tra_1}- \sum_{\tra_0}\le(\kert_{\text{ER}}\kert^{-1}_{\text{RR}}\ri)_{\tes}^{\ \ \ \tra_0}A_{\tra_0 \tra_1}\ri]\, \nonumber
\end{align}
is the mean prediction. Equations~(\ref{NNNGPPdash}) and~(\ref{NNNGPPdashdash}) with $\overline{\phi}_{i}^{\ \tra}=\sum_{\tra'}\le(\kert_{\text{RR}}^{-1}\ri)^{\tra \tra'}\le(y_{\text{R}}\ri)_{i;\tra'}$ are actionable, i.e., easy to program.

\end{document}